\documentclass[12pt]{article} 
\fontsize{12}{14}\selectfont
\usepackage{amsmath}
\usepackage{times}
\usepackage{graphicx}
\usepackage{color}
\usepackage{tikz}
\usetikzlibrary{backgrounds,positioning,calc,shapes,arrows.meta,arrows,decorations.markings}
\usepackage{listofitems}
\usepackage{multirow}
\usepackage{subcaption}
\usepackage{pifont}
\usepackage{amssymb}
\usepackage{amsfonts}
\usepackage{float}
\usepackage{pdflscape}
\usepackage{array}
\usepackage{algorithm}
\usepackage{algcompatible}
\usepackage{algpseudocode}
\algdef{SE}[SUBALG]{Indent}{EndIndent}{}{\algorithmicend\ }%
\algtext*{Indent}
\algtext*{EndIndent}
\usepackage{afterpage}
\usepackage{caption}

\makeatletter
\renewcommand\footnotesize{%
   \@setfontsize\footnotesize\@xiipt{14}%
   \abovedisplayskip 10\p@ \@plus2\p@ \@minus5\p@
   \abovedisplayshortskip \z@ \@plus3\p@
   \belowdisplayshortskip 6\p@ \@plus3\p@ \@minus3\p@
   \def\@listi{\leftmargin\leftmargini
               \topsep 6\p@ \@plus2\p@ \@minus2\p@
               \parsep 3\p@ \@plus2\p@ \@minus\p@
               \itemsep \parsep}%
   \belowdisplayskip \abovedisplayskip
}
\makeatother

\usepackage[authoryear]{natbib}
\usepackage{rotating}
\usepackage{bbm}
\usepackage{latexsym}

\newcommand{\calculateYShift}[1]{
  \pgfmathsetmacro{\yshift}{(#1 - 1) * 3 + min(max(#1 - 1, 0), .15)}
}

\newcommand{\captionfonts}{\normalsize}

\makeatletter  
\long\def\@makecaption#1#2{%
  \vskip\abovecaptionskip
  \sbox\@tempboxa{{\captionfonts #1: #2}}%
  \ifdim \wd\@tempboxa >\hsize
    {\captionfonts #1: #2\par}
  \else
    \hbox to\hsize{\hfil\box\@tempboxa\hfil}%
  \fi
  \vskip\belowcaptionskip}
\makeatother   

\newcommand{\thisnode}[3]{#1#2#3}
\newcommand{\x}[1]{\the\numexpr#1\relax}

\newcommand{\currentpoint}{(0,0)} 

\newcommand{\startline}[2]{
    \coordinate (#2) at #1;
    \renewcommand{\currentpoint}{#1}
}

\newcommand{\straightline}[3][]{
    \coordinate (#3) at #2;
    \draw[#1] \currentpoint -- #2;
    \renewcommand{\currentpoint}{#2}
}

\newcommand{\bezierline}[5][]{
    \coordinate (#5) at #2;
    \draw[#1] \currentpoint .. controls #3 and #4 .. #2;
    \renewcommand{\currentpoint}{#2}
}

\title{\noindent Intrinsic Rewards for Exploration without Harm \\ from Observational Noise: A Simulation Study \\ Based on the Free Energy Principle}
\author{\noindent Theodore Jerome Tinker, Kenji Doya, Jun Tani$^*$}

\makeatletter
\def\@maketitle{
    \vspace*{7cm}
    {\LARGE \@title} \\
    \vskip .4em
    {\bf \@author} \\
    \noindent{\bf Keywords:} Reinforcement Learning, The Free Energy Principle, \\ Active Inference, Entropy, Curiosity
    \par
    \vskip 1.2em
    {\fontsize{12}{14}\selectfont 
    \indent\indent\indent\indent Cognitive Neurorobotics Research Unit, Okinawa Institute of Science \\[-1ex]
    \indent\indent\indent\indent and Technology Graduate University  \\[-1ex] 
    \indent\indent\indent\indent 1919-1 Tancha, Onna-san 904-0495, Okinawa, Japan \\[-1ex]
    \indent\indent\indent\indent $^*$ \ \ Correspondence: jun.tani@oist.jp \\[-1ex]
    }
    \newline
    \vskip 1.5em
}
\makeatother

\begin{document}

\maketitle

%


%
\newpage
\begin{center} {\bf Abstract} \end{center}

In Reinforcement Learning (RL), artificial agents are trained to maximize numerical rewards by performing tasks. Exploration is essential in RL because agents must discover information before exploiting it. Two rewards encouraging efficient exploration are the entropy of action policy and curiosity for information gain. Entropy is well-established in literature, promoting randomized action selection. Curiosity is defined in a broad variety of ways in literature, promoting discovery of novel experiences. One example, prediction error curiosity, rewards agents for discovering observations they cannot accurately predict. However, such agents may be distracted by unpredictable observational noises known as curiosity traps. Based on the Free Energy Principle (FEP), this paper proposes hidden state curiosity, which rewards agents by the KL divergence between the predictive prior and posterior probabilities of latent variables. We trained six types of agents to navigate mazes: baseline agents without rewards for entropy or curiosity, and agents rewarded for entropy and/or either prediction error curiosity or hidden state curiosity. We find entropy and curiosity result in efficient exploration, especially both employed together. Notably, agents with hidden state curiosity demonstrate resilience against curiosity traps, which hinder agents with prediction error curiosity. This suggests implementing the FEP may enhance the robustness and generalization of RL models, potentially aligning the learning processes of artificial and biological agents.


\section{Introduction}
\label{chap:introduction}

Reinforcement Learning (RL) is a machine learning algorithm for training artificial agents to perform tasks by awarding or punishing their actions with numerical rewards \citep{actor_critic_2, DQN_1, DQN_2}. This can be interpreted as akin to biological agents learning through evolution. Extrinsic rewards are awarded at human discretion based on the tasks at hand. Intrinsic rewards are generated by agents themselves (with human-provided hyperparameters) based on other goals such as exploration. Exploration is an important but difficult aspect of RL because an agent can only exploit knowledge after learning that knowledge. An exploration phase may be implemented just by selecting random actions for the agent, but this can be inefficient, especially with high-dimensional continuous state-action spaces and sparse extrinsic rewards. Hence, we study two intrinsic rewards for efficient exploration: entropy in the action-space for control as inference \citep{cai_vs_aif} and curiosity about the environment for active inference \citep{tschantz2020reinforcement, tschantz2023scaling}. Meanwhile, Friston's Free Energy Principle (FEP) describes biological neuroscience with Bayesian statistics applicable to machine learning and AI \citep{kaplan2018planning, parr_friston_2019}. Our goal is to share a novel definition of curiosity derived from the FEP and demonstrate its robust usefulness in exploration.

Curiosity in RL has been presented in many ways, typically based on an agent's ability to predict future observations. This leverages an adversarial relationship between the agent’s predictive accuracy using a forward or generative model, also known as a transitioner or world model, and its pursuit of observations that challenge this model's accuracy. For example, \citet{schmidhuber2010formal} critiqued that curiosity gauged by ``mean squared prediction error or similar measures,'' which \citet{oudeyer2007intrinsic} called Predictive Novelty Motivation and we call prediction error curiosity, ``may fail whenever high prediction errors do not imply expected prediction progress, e.g., in noisy environments.'' Alternative forms of curiosity from these sources included estimating likelihoods of events, measuring prediction errors probabilistically, or assessing improvement in predictions, but these methods can have great computational costs. Moreover, Oudeyer and Kaplan note ``in certain application contexts... intrinsic openness is a weakness'' and counterproductive. 

\citet{trap} elaborated on Schmidhuber's critique of prediction error curiosity in noisy environments, which may be an example of what Oudeyer and Kaplan called a weakness of intrinsic openness. Pathak et al. asked readers to consider an agent which could observe tree leaves randomly dancing in the wind. The agent's forward model would never be able to perfectly predict such observations, so the agent might become fixated on these leaves like a moth attracted to a lamp. Thus, such observational noises or expected uncertainties are called curiosity traps. To remedy this, Pathak et al. trained an inverse dynamics model to predict the agent's action between two consecutive observations, thereby encoding observations into a latent space without noisy details, relevant only to the agent's actions. Then, the agent's forward model could be trained to predict these refined latent states instead of chaotic observations. Prediction error curiosity based on that forward model could ignore the environment's irrelevant noise and thus be minimally impacted by curiosity traps, instead focusing on unexpected uncertainties. However, this could not function well with rarely experienced interactions. Pathak et al. suggested storing events in a memory buffer for experience replay, which we implement.

\citet{schwartenbeck2019} derived two intrinsic rewards for exploration directly from the FEP. An agent with these intrinsic rewards uses Bayesian inference to minimize free energy, meaning not only maximizing extrinsic rewards but also developing a thorough understanding of the environment. The first intrinsic reward, parameter exploration, encourages active learning: the agent seeks to resolve uncertainty about how its actions are rewarded. The second intrinsic reward, hidden state exploration, encourages active inference: the agent seeks to take actions which reveal uncertain observations. Together these intrinsic rewards establish curiosity for both the environment and the task at hand. However, Schwartenbeck et al. assumed agents in T-mazes knew the mazes had two arms, the left providing a constant extrinsic reward, the right providing an uncertain extrinsic reward. Agents in RL typically must infer such knowledge from observations, so these intrinsic rewards are not easily applied to RL directly.

\citet{kawahara} derived both entropy and an RL-applicable definition of curiosity from the FEP. They constructed a forward model as a Bayesian Neural Network (BNN) such that the weight-parameters are not fixed, but are drawn from a multivariate Gaussian distribution using the reparameterization trick for stochastic optimization \citep{bnn, kingma2015variational}. Kawahara et al. gauged curiosity values based on Kullback–Leibler divergence comparing the weights' distribution before and after learning to predict each new observation with minimal free energy. This probabilistic approach must consider uncertainty in observational noise, so like the curiosity of Pathak et al., this curiosity should be able to effectively explore without negative influence from curiosity traps. However, a BNN's computational cost is high, and Kawahara et al. only worked in terms of a Markov Decision Process (MDP) where states are completely observed. Additionally, this method produces just a single curiosity value per training update, which, if applied to an entire batch, would only produce one curiosity value for the batch as a whole, overlooking the contributions of each observation individually. Identifying which parts of the batch are important for exploration and which are not is a computationally demanding task because each observation must be evaluated one at a time. 

In this paper, we overcome the problem of prediction error curiosity using hidden state curiosity defined in section \ref{sec:pvrnn}. Like the curiosity of Kawahara et al., hidden state curiosity is derived from the FEP and gauged by the Kullback-Leibler divergence between predictive prior and posterior over future states (under a particular policy). Like the curiosity of Pathak et al., those states are efficiently encoded as latent variables. We train six types of agents: a baseline with no intrinsic rewards, entropy-driven, prediction error curious, hidden state curious, and two hybrids combining entropy with each form of curiosity. The abilities of these agents to find goals in a biased T-maze or an expanding T-maze (first a T-maze, then a double T-maze, and then a triple T-maze) will test the following two hypotheses:

\begin{enumerate}
\item Entropy and curiosity improve agent exploration, especially when both are implemented together as implied by the FEP. 
\item Prediction error curiosity can be negatively influenced by observational noise also known as curiosity traps, while hidden state curiosity can be more resilient to such curiosity traps.
\end{enumerate}

\noindent 
The results in section \ref{chap:experiments} are presented to evaluate the validity of these hypotheses, contributing to our understanding of artificial intelligence exploration in complex environments.
\section{Prior Studies}
\label{chap:prior}

\subsection{Reinforcement Learning}
\label{sec:RL}

In Reinforcement Learning (RL), an agent experiences an episode as a sequence of transitions of the form $\{ o_t,$ $a_t,$ $r_t,$ $o_{t+1},$ $done_t \}$. Variables $o_t$ and $o_{t+1}$ are the agent's observations at times $t$ and $t+1$ equal to (in a Markov Decision Process, MDP) or derived from (in a Partially Observable Markov Decision Process, POMDP) the complete environmental states $s_t$ and $s_{t+1}$. Variable $a_t$ is the agent's action performed at time $t$. Variable $r_t$ is the extrinsic reward the agent obtained by performing that action. And variable $done_t$ is 1 if time $t$ was the final step in the episode or 0 otherwise. 

In the Actor-Critic method \citep{actor_critic_2}, an agent has at least two neural networks using parameters $\phi$ and $\theta$ to instantiate implicit  (i.e., amortise) mappings: an actor network $\pi_\phi(o_t) \rightarrow a_t$ also known as a policy, which chooses actions based on observations, and a critic network $Q_\theta (o_t, a_t) \rightarrow \widehat{Q}_t$, which predicts future rewards based on observations and actions to estimate the state-action value function. The critic's target value $Q_t$ is $r_t$ plus future rewards through bootstrapping (see equation \ref{eq:target_value}, similar to Bellman's equation) predicted by a target critic $Q_{\bar{\theta}}$. The target critic begins with parameters starting equal to the critic's, then slowly learns alongside the critic with Polyax averaging such that $\bar{\theta} \leftarrow \tau\theta + (1 - \tau) \bar{\theta}$ with hyperparameter $\tau \in [0,1]$ being the soft update coefficient. 

An ensemble of multiple critics (and an equal number of target critics) can be implemented, in which case the actor's loss function (equation \ref{eq:actor_with_entropy}) utilizes the lowest predicted value among all critics. Furthermore, the actor's training can be staggered with a delay of $d$ epochs relative to the critics' training, ensuring less frequent actor training for stability. Regardless of these implementations, the agent's actor and critics shall be trained off-policy with experience replay by randomly sampling a batch of transitions from a memory buffer.

When using the Soft-Actor-Critic method (SAC) to implement entropy \citep{SAC}, the agent's actions are randomly sampled with the reparameterization trick for stochastic optimization \citep{kingma2015variational}. The actor generates two vectors the size of an action: mean $\mu_t$, made without activation, and standard deviation $\sigma_t$, made with softplus activation. The action is finally selected by combining these vectors with noise $\epsilon_t$ sampled from the Unit Gaussian distribution $N(0,1)$, resulting in $a_t = \tanh(\mu_t + \sigma_t \circ \epsilon_t)$. Entropy in action-choice is intrinsically rewarded in the actor's loss function which trains the actor to minimize $J_\pi$:

\begin{equation}\label{eq:actor_with_entropy}
J_\pi(\phi) = \mathbb{E}_{o_t \sim D, a_t \sim \pi_\phi} \left[ - Q_\theta (o_t, a_t)\right] - \alpha \mathcal{H}(\pi_{\phi}(a_{t} | o_{t}))
\end{equation}

\noindent 
with $D$ being the distribution of observations in the environment, non-negative hyperparameter $\alpha$ being the relative importance of entropy, and $\mathcal{H}(\pi_{\phi} (a_{t} | o_{t}))$ being the self-information (i.e., negative log probability) of an action $a_t$ given observation $o_t$, whose expectation can be read as an action entropy. The hyperparameter $\alpha$ can be chosen manually, or it can be a dynamic parameter which is trained based on another hyperparameter, target entropy $\bar{\mathcal{H}}$. In that case, $\alpha$ is chosen to minimize $J_\alpha$ in equation \ref{eq:alpha}. If the difference between entropy and target entropy is negative, actions are too probable, so $\log(\alpha)$ will be increased to encourage more entropy; if the difference is positive, actions are too improbable, so $\log(\alpha)$ will be decreased to encourage less entropy.

\begin{equation}\label{eq:alpha}
J_\alpha(\alpha) = \log (\alpha) \cdot (\mathcal{H}(\pi(a_t|o_t)) - \bar{\mathcal{H}})
\end{equation}

Meanwhile, the SAC's critic is trained to minimize $J_Q$ in the following target value and loss function. We here also include an intrinsic reward for curiosity, $P_t$, which can take several forms. In subsequent sections we will consider the functional form of these alternatives and, in the final section, use numerical experiments to quantify their contribution to optimal behavior, as assessed via reward learning.

\begin{align}\label{eq:target_value}
Q(t) &= r_t + \eta P_t + \nonumber \\ 
&\gamma (1 - done_t) \mathbb{E}_{o_{t+1} \sim D, a_{t+1} \sim \pi_\phi} [Q_{\bar{\theta}}(o_{t+1}, a_{t+1})] - \alpha \mathcal{H}(\pi_{\phi} (a_{t+1} | o_{t+1}))
\end{align}

\begin{equation}\label{eq:critic_func}
J_Q(\theta) = \mathbb{E}_{o_t \sim D, a_{t} \sim \pi_\phi} [\left(Q_\theta(o_t, a_t) - Q(t)\right)^2]
\end{equation}

\noindent
with non-negative hyperparameter $\eta$ being the relative importance of curiosity value $P_t$ and hyperparameter $\gamma \in [0,1]$ being the bootstrapping discount rate.

\subsection{Recurrent RL}
\label{sec:RNN}

\begin{figure}[ht]
\centering

\begin{subfigure}[b]{\textwidth}
\centering
\begin{tikzpicture}[>=stealth,font=\sffamily]

    \node[draw,fill=gray!10,rectangle,minimum width=5cm,minimum height=5cm] (box) at (0,0) {};
    \node[above] at (box.north) {Actor (RNN)};

    \node[draw,fill=white,rectangle,minimum size=1cm] (ha_t) at (-1.2cm,-.35) {$h^\phi_{t-1}$};
    \node[draw,fill=white,rectangle,minimum size=1cm,above=.2cm of ha_t] (a_t) {$a_{t-1}$};
    \node[draw,fill=white,rectangle,minimum size=1cm,right=.2cm of a_t] (o_tp1) {$o_{t}$};
    \node[draw,fill=white,rectangle,minimum size=1cm] (ha_tp1) at ($(ha_t -| o_tp1) + (1.2,0)$) {$h^\phi_{t}$};
    \node[draw,fill=white,rectangle,minimum size=1cm,below=.2cm of ha_tp1] (a_tp1) {$a_{t}$};
    \node[draw,fill=white,rectangle,minimum size=1cm, left=.5cm of a_tp1] (loss) {Equation \ref{eq:actor_with_entropy}};
    \draw[->] (ha_t) + (-1,1) -- (ha_t);
    \draw[->] (ha_t) + (-1,.5) -- (ha_t);
    \draw[->] (ha_t) + (-1,0) -- (ha_t);
    \draw[->] (ha_t) -- (a_t);
    \draw[->] (ha_t) -- (ha_tp1);
    \draw[->] (a_t) .. controls +(.75,-1) and +(-1,.5) .. node[midway, above, sloped] {} (ha_tp1);
    \draw[->,double=white,color=black] (a_t) .. controls +(0,1) and +(0,1) .. node[midway, above, sloped] {ENV} (o_tp1);
    \draw[->] (o_tp1) -- (ha_tp1);
    \draw[->,red] (loss) -- (a_tp1);
    \draw[->] (ha_tp1) -- ($(ha_tp1) + (1,0)$);
    \draw[->] (ha_tp1) -- (a_tp1);
\end{tikzpicture}
\hspace*{1cm}
\begin{tikzpicture}[>=stealth,font=\sffamily]

    \node[draw,fill=gray!10,rectangle,minimum width=5cm,minimum height=4.5cm] (box) at (0,0) {};
    \node[above] at (box.north) {Critic (RNN)};

    \node[draw,fill=white,rectangle,minimum size=1cm] (hc_t) at (-1.2,0) {$h^\theta_{t-1}$};
    \node[draw,fill=white,rectangle,minimum size=1cm,above=.2cm of hc_t] (a_tp1) {$a_{t}$};
    \node[draw,fill=white,rectangle,minimum size=1cm,right=.2cm of a_tp1] (o_tp1) {$o_{t}$};
    \node[draw,fill=white,rectangle,minimum size=1cm,below right=.2cm and .2cm of o_tp1] (hc_tp1) {$h^\theta_{t}$};
    \node[draw,fill=white,rectangle,minimum size=1cm,below=.2cm of hc_tp1] (Q) {Q};
    \node[draw,fill=white,rectangle,minimum size=1cm,left=.5cm of Q] (target) {Equation \ref{eq:critic_func}};

    \draw[->] (hc_t) + (-1,1) -- (hc_t);
    \draw[->] (hc_t) + (-1,.5) -- (hc_t);
    \draw[->] (hc_t) + (-1,0) -- (hc_t);
    \draw[->] (hc_t) -- (hc_tp1);
    \draw[->] (a_tp1) .. controls +(.75,-1) and +(-1,.5) .. node[midway, above, sloped] {} (hc_tp1);
    \draw[->] (o_tp1) -- (hc_tp1);
    \draw[->] (hc_tp1) -- (Q);
    \draw[<-,red] (Q) -- (target);
    \draw[->] (hc_tp1) -- ($(hc_tp1) + (1,0)$);
\end{tikzpicture}
\caption{}
\label{fig:yes_recur}
\end{subfigure}

\begin{subfigure}[b]{\textwidth}
\centering
\begin{tikzpicture}[>=stealth,font=\sffamily]

    \node[draw,fill=gray!10,rectangle,minimum width=4.25cm,minimum height=3cm] (box) at (0,0) {};
    \node[above] at (box.north) {Actor (Forward's Hidden State)};

    \node[draw,fill=white,rectangle,minimum size=1cm] (ha_tp1) at (1.4,.6) {$h^q_{t}$};
    \node[draw,fill=white,rectangle,minimum size=1cm,below=.2cm of ha_tp1] (a_tp1) {$a_{t}$};
    \node[draw, fill=white,rectangle, minimum size=1cm, left=.5cm of a_tp1] (loss) {Equation \ref{eq:actor_with_entropy}};
    \draw[->,red] (loss) -- (a_tp1);

    \draw[->] (ha_tp1) -- (a_tp1);
    
\end{tikzpicture}
\hspace*{1cm}
\begin{tikzpicture}[>=stealth,font=\sffamily]

    \node[draw,fill=gray!10,rectangle,minimum width=4.25cm,minimum height=3cm] (box) at (0,0) {};
    \node[above] at (box.north) {Critic (Forward's Hidden State)};

    \node[draw,fill=white,rectangle,minimum size=1cm] (hc_tp1) at (1.4cm, .6cm) {$h^q_{t}$};
    \node[draw,fill=white,rectangle,minimum size=1cm,left=.2cm of hc_tp1] (a_t) {$a_{t}$};
    \node[draw,fill=white,rectangle,minimum size=1cm,below=.2cm of hc_tp1] (Q) {Q};
    \node[draw,fill=white,rectangle,minimum size=1cm,left=.5cm of Q] (target) {Equation \ref{eq:critic_func}};

    \draw[->] (hc_tp1) -- (Q);
    \draw[->] (a_t) -- (Q);
    \draw[<-,red] (Q) -- (target);
    
\end{tikzpicture}
\caption{}
\label{fig:no_recur}
\end{subfigure}

\caption{(a) Implementing recurrent layers in an actor model and a critic model. Notice the previous action $a_{t-1}$ is implicitly included in $h^\theta_{t-1}$. (b) Implementing the forward model's hidden state $h^q$ in an actor model and a critic model. Black arrows indicate forward computations. Red arrows indicate loss functions for backpropagation.}
\label{fig:recur}
\end{figure}
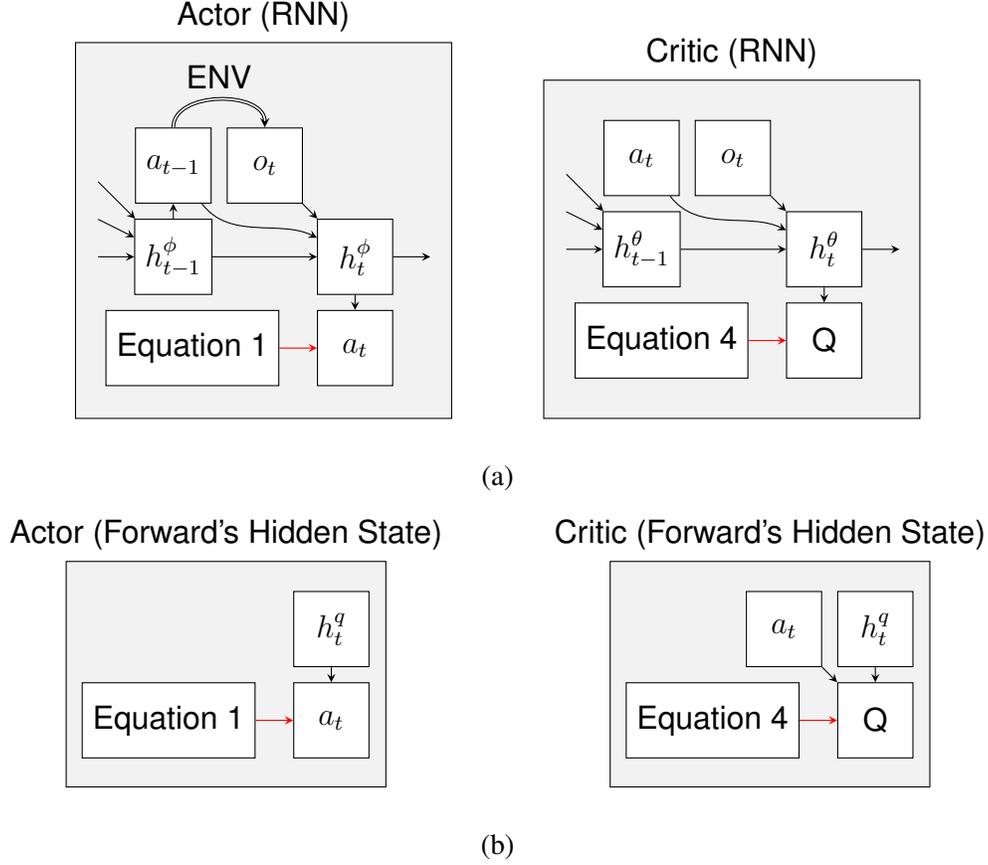

\citet{recurrent} brought recurrent neural networks (RNN) to RL by providing the actor and critic recurrent layers utilizing hidden states $h^\phi_t$ and $h^\theta_t$ respectively (see figure \ref{fig:yes_recur}). Such an actor or critic requires two additional inputs: the previous action $a_{t-1}$ and the previous hidden state $h^\phi_{t-1}$ or $h^\theta_{t-1}$ (in the critic's case, $h^\theta_{t-1}$ contains $a_{t-1}$ implicitly), with $a_{-1}$, $h^\phi_{-1}$, and $h^\theta_{-1}$ being initialized as zero-filled tensors. The actor's hidden states (and, symmetrically, the critic's hidden states) would be calculated as 

$$h^\phi_1, ..., h^\phi_{t} = \text{RNN}_\phi (o_1||a_0, ..., o_{t}||a_{t-1})$$

\noindent
with $||$ denoting concatenation. To train recurrent models with experience replay, a recurrent memory buffer must contain whole episodes beginning to end for temporal context. Episodes may vary in length, so batches sampled from recurrent memory buffers have all episodes standardized to the maximum episode's length using zero-filled transitions. Therefore, transitions of the batch include another variable, $mask_t$, which is 1 if the transition was actually within the episode or 0 if the transition was added for standard length.

We utilize this method and another method described in section \ref{sec:pvrnn} depicted in figure \ref{fig:no_recur}. Using either of these methods enables the agent to maintain a continuity of experience, thus giving the agent a temporal edge in learning and decision-making. This may be essential for navigating environments where states are not directly observable.

\subsection{Curiosity Derived from the FEP}

Among various definitions of curiosity such as those from \citet{oudeyer2007intrinsic}, \citet{schmidhuber2010formal}, \citet{trap}, and \citet{schwartenbeck2019}, the most relevant to this paper is that of \citet{kawahara} because it is directly derived from the FEP for RL. Here we first describe the FEP as utilized by Kawahara et al., then describe their definition of curiosity in that framework. 

The cornerstone of the FEP is Bayes' theorem, which describes the relationship between the prior probability distribution of parameter $X$, $p(X)$, and the posterior probability distribution given data $D$, $p(X|D)$, expressed as

$$p(X|D) = \frac{p(D|X)p(X)}{p(D)} = \frac{p(D|X)p(X)}{\int p(D,X) dX}.$$

\noindent 
However, $\int p(D,X) dX$ may be far too computationally expensive to calculate. Instead we can approximate $p(X|D)$ with the predictive distribution $q(X)$. The difference between $q(X)$ and $p(X|D)$ can be measured using Kullback–Leibler divergence:

\begin{align}
D_{KL} [q(X) || p(X|D)] &= \int q(X)\log\frac{q(X)}{p(X|D)} dX \nonumber \\
&= \int q(X)\log\frac{q(X)p(D)}{p(D,X)} dX \nonumber \\
&= \int q(X)\log\frac{q(X)p(D)}{p(X)p(D|X)} dX.
\end{align}

\noindent
Because that integral is in terms of $X$ and $p(D)$ is independent of $X$, we may remove $p(D)$ from the integral with the laws of logarithms. This reveals the concept of variational free energy $F$, defined as

\begin{align}
D_{KL} [q(X) || p(X|D)] &= \underbrace{D_{KL} [q(X) || p(X)]}_{\text{complexity}} - \underbrace{\mathbb{E}_{X} [\text{log} p(D|X)]}_{\text{accuracy}} + \text{ log } p(D) \nonumber \\
&= F + \log p(D). 
\end{align}

\noindent
With this, we may estimate $q(X)$ as $\underset{q(X)}{\text{arg min }} F.$ 

Given time $\tau \geq t+1$, expected free energy $G_\tau$ is equal to free energy $F$ multiplied by likelihood $P(D_\tau|X_\tau)$, such that

\begin{align}\label{eq:this_one}
G_\tau &= \mathbb{E}_{p(D_\tau | X_\tau)} [F] \nonumber \\
&= \mathbb{E}_{p(D_\tau | X_\tau)} [\int q(X_\tau) \log \frac{q(X_\tau)}{p(D_\tau,X_\tau)} dX] \nonumber \\
&= \mathbb{E}_{p(D_\tau|X_\tau)} [\mathbb{E}_{q(X_\tau)} [\log \frac{q(X_\tau)}{p(X_\tau|D_\tau)} - \log p(D_\tau)]].
\end{align}
\noindent 
Because $q$ approximates $p$ and $q(D_\tau|X_\tau) q(D_\tau) = q(D_\tau,X_\tau)$,

\begin{align}
G_\tau &\approx \mathbb{E}_{q(D_\tau,X_\tau)} [\log \frac{q(X_\tau)}{q(X_\tau|D_\tau)} - \log p(D_\tau)] \nonumber \\
&= - \mathbb{E}_{q(D_\tau, X_\tau)} [\log \frac{q(X_\tau|D_\tau)}{q(X_\tau)}] - \mathbb{E}_{q(D_\tau)} [\log p(D_\tau)] \nonumber \\
&= - \underbrace{\mathbb{E}_{q(D_\tau)} [\overbrace{D_{KL} [q(X_\tau|D_\tau) || q(X_\tau)]}^{\text{Bayesian Surprise}}]}_{\text{Epistemic Value or Mutual Information}} - \underbrace{\mathbb{E}_{q(D_\tau)} [\log p(D_\tau)]}_{\text{Extrinsic Value}}. 
\end{align}

\noindent 
\citet{kawahara} described how this equation can be used to interpret biological agents' behaviors. The Extrinsic Value term represents the biological agents' prior preferences or desires, much like extrinsic reward $r_t$. The Epistemic Value term represents how living creatures choose actions to find data $D$ providing information about parameter $X$, reducing uncertainty. Hence, biological agents seek to satisfy their desires while also acquiring unknown information. 

Kawahara et al. applied the FEP to RL by training agents to choose actions which minimize expected free energy, emulating biological thought. Let $D_\tau = \{s_\tau, a_\tau\}$ (with the understanding that Kawahara et al. considered states to be completely observed such that $o_\tau = s_\tau$) so expected free energy can be rewritten as 

\begin{align} \label{eq:applying_D}
G_\tau &= -\mathbb{E}_{q(s_\tau, a_\tau, X_\tau)} [\log \frac{p(X_\tau|s_\tau,a_\tau)}{q(X_\tau)}] - \mathbb{E}_{q(s_\tau, a_\tau)} [\log p(s_\tau, a_\tau)] \nonumber \\
&= -\mathbb{E}_{q(s_\tau, a_\tau, X_\tau)} [\log \frac{p(X_\tau,a_\tau|s_\tau)}{q(X_\tau)p(a_\tau|s_\tau)}] - \mathbb{E}_{q(s_\tau, a_\tau)} [\log p(s_\tau, a_\tau)] \nonumber \\
&\approx -\mathbb{E}_{q(s_\tau, a_\tau, X_\tau)} [\log \frac{q(X_\tau|s_\tau)q(a_\tau|s_\tau,X_\tau)}{q(X_\tau)p(a_\tau|s_\tau)}] - \mathbb{E}_{q(s_\tau, a_\tau)} [\log p(s_\tau, a_\tau)] \nonumber \\
&= -\mathbb{E}_{q(a_\tau | s_\tau, X_\tau)q(s_\tau)} [D_{KL}[q(X_\tau|s_\tau)||q(X_\tau)]] \nonumber \\
&- \mathbb{E}_{q(s_\tau,X_\tau)} [D_{KL}[q(a_\tau|s_\tau,X_\tau)||p(a_\tau|s_\tau)]] \nonumber \\
&- \mathbb{E}_{q(s_\tau,a_\tau)}[\log p(s_\tau,a_\tau)]. 
\end{align}

\noindent
Consider now a forward model $f_w$ which is a Bayesian Neural Network (BNN) \citep{bnn}. Because this forward model predicts the next state given the current state and action, its weight parameters $w_\tau$ can be interpreted as the agent's latent understanding of the environment; thus, let $X_\tau = w_\tau$ such that the first term of equation \ref{eq:applying_D} reflects how the agent's understanding of the environment changes based on experiencing a state. Let $q_\psi = \mathcal{N}(w_\tau|\mu,\sigma)$ with $\psi = \{\mu,\sigma\}$, so a Soft Actor Critic actor $\pi_\phi$ can be trained to approximate $\pi_\phi(a_\tau|s_\tau) \approx q(a_\tau|s_\tau,w_\tau)$. This allows rewriting expected free energy as 

\begin{align} \label{eq:replaced_X}
G(s_\tau, a_\tau) &= -\mathbb{E}_{q(a_\tau | s_\tau, w_\tau)q(s_\tau)} [D_{KL}[q(w_\tau|s_\tau)||q(w_\tau)]] \nonumber \\
&- \mathbb{E}_{q(s_\tau,a_\tau)} [D_{KL}[\pi_{\phi}(a_\tau|s_\tau)||p(a_\tau|s_\tau)]] \nonumber \\
&- \mathbb{E}_{q(s_\tau,a_\tau)}[\log p(s_\tau,a_\tau)].
\end{align}

\noindent
Finally, let $\tau = t$, let prior preference extrinsic value $\log p(s_t, a_t)$ be extrinsic reward $r(s_t, a_t) = r_t$, and recall the forward model's objective is predicting $s_{t+1}$, so expected free energy can be rewritten as 

\begin{align}\label{eq:all_intrinsic}
G(s_t, a_t) &= -D_{KL}[q_\psi(w_t|s_{t+1})||q_\psi(w_t)] - \log p(s_t,a_t) \nonumber \\
&- D_{KL}[\pi_\phi(a_t|s_t)||p(a_t|s_t)] \nonumber \\
&= -D_{KL}[q_\psi(w_t|s_{t+1})||q_\psi(w_t)] - \log p(s_t,a_t) \nonumber \\
&- \int \pi_\phi(a_t|s_t) \log \pi_\phi(a_t|s_t) da_t + \int \pi_\phi(a_t|s_t) \log p(a_t|s_t) da_t\nonumber \\
&= -\underbrace{D_{KL}[q_\psi(w_t|s_{t+1})||q_\psi(w_t)]}_{\text{Curiosity}} - \underbrace{r(s_t, a_t)}_{\text{Extrinsic Reward}} \nonumber \\
&- \underbrace{\mathcal{H}(\pi_\phi(a_t|s_t))}_{\text{Entropy}} - \underbrace{\mathbb{E}_{\pi_\phi(a_t|s_t)} [\log p(a_t|s_t)]}_{\text{Imitation}}.
\end{align}

\noindent 
In this formation, the RL implementation of the FEP presents extrinsic rewards, entropy, and curiosity. (Imitation, an intrinsic reward which we do not explore in this paper, could further be used to train an agent to emulate the behavior of another agent.) Equations \ref{eq:actor_with_entropy} and \ref{eq:target_value} can incorporate these entropy and curiosity values or approximations thereof, multiplied by non-negative hyperparameters $\alpha$ and $\eta$. Calculating the curiosity value $P_t = D_{KL}[q_\psi(w_t|s_{t+1})||q_\psi(w_t)]$ involves two distributions: $q_\psi(w_t)$, the distribution of $f_w$'s weights, and $q_\psi(w_t|s_{t+1})$, the same distribution after considering the next state. The first distribution, $q_\psi(w_t)$, is readily available; it may be called a predictive prior. We can then obtain the predictive posterior distribution $q_\psi(w_t|s_{t+1})$ as if training the forward model to predict $s_{t+1}$ with minimal free energy $F$, calculated as

\begin{equation}\label{eq:free_energy}
F = \underbrace{D_{KL}[q_\psi(w_t)||p_\psi(w_t)]}_{\text{Complexity}} - \underbrace{\mathbb{E}_{q_\psi(w_t)}[\log p_\psi(s_{t+1}|w_t)]}_{\text{Accuracy}}.
\end{equation}

\noindent 
In practice, prior distribution $p_\psi(w_t)$ can be the unit Gaussian $\mathcal{N}(0,1)$, and the complexity term is multiplied by a non-negative hyperparameter $\beta$ describing its relative importance to accuracy. 

Thusly, Kawahara et al. used the FEP's framework to define curiosity such that an observation has a low curiosity value if the forward model does not need to change much to accommodate it, or a high curiosity value if the forward model must change drastically. This encourages an adversarial relationship between the agent's forward model and actor: the forward model trains to improve its weights representing a probabilistic interpretation of the environment, but the critic rewards the actor for finding information which substantially alters the forward model's weights. This active inference complements entropy's control as inference. Importantly, the agent's probabilistic interpretation of the environment should account for observational noise, so observing anticipated noise should not alter it much; hence, the free energy based curiosity defined by Kawahara et al. should be able to effectively explore without negative influence from curiosity traps. However, this definition of curiosity is constrained regarding batch processing: individual transitions within a batch may differ in exploratory importance, but comparing the forward model's weights before and after training with the entire batch as a whole returns only one curiosity value. Investigating the significance of each transition individually requires great computational cost. That limitation, and the restriction to fully observable MDP, suggest there are opportunities for further development. 
\section{Proposed Model}
\label{sec:pvrnn}

Motivated by Kawahara et al. and Pathak et al., we define hidden state curiosity using a forward model with the architecture of a Variational RNN (VRNN) \citep{chung2016recurrent}. Our forward model $f_\psi$, pictured in figure \ref{fig:forward} and described by algorithm \ref{algo:forward}, is recurrent using hidden state $h^q_t$, enabling the accounting of temporal dependencies and uncertainties in a Partially Observable Markov Decision Process (POMDP). This provides a second method for providing the actor and critic temporal knowledge: replacing observations in the models' inputs with $h^q_t$ (see figure \ref{fig:no_recur}). We apply this method to the actor, with a training-delay of $d=2$, while two critics use their own recurrent layers, as shown in figure \ref{fig:yes_recur} (see section \ref{sec:RL} regarding delayed actor training and multiple critics). Unlike the Bayesian Neural Network (BNN) used by Kawahara et al., our forward model does not have probabilistic weights; instead, we use the reparameterization trick in the style of a SAC actor or Variational Bayes Autoencoder (VAE) to sample prior and posterior inner states $z^p_t$ and $z^q_t$ from corresponding probability distributions $p(z_t)=\mathcal{N}(\mu^p_t,\sigma^p_t)$ and $q(z_t)=\mathcal{N}(\mu^q_t,\sigma^q_t)$. These distributions are derived from the previous hidden state $h^q_{t-1}$, the previous action $a_{t-1}$, and in the case of the posterior inner state, the current observation $o_t$. Posterior inner state $z^q_t$ and previous hidden state $h^q_{t-1}$ are used to generate hidden state $h^q_t$.

\input{tikz/forward.tex}

This forward model trains to generate accurate predictions $\widehat{o}_{t+1}$ of upcoming observations $o_{t+1}$ with hidden states $h^q_t$ and actions $a_t$ while minimizing the Kullback-Leibler divergence comparing $p(z_t)$ and $q(z_t)$. The relationship between $p(z_t)$ and $q(z_t)$ is parallel to the relationship between Kawahara et al.'s $q(w_t)$ and $q(w_t|s_t)$, used in equation \ref{eq:replaced_X} to replace $q(X_t)$ and $q(X_t|s_t)$ in equation \ref{eq:applying_D}: these pairs are an agent's probabilistic understanding of its environment before and after experiencing a state or observation. Thus, this forward model is trained to minimize free energy $F$ with equation \ref{eq:free_energy} rewritten as equation \ref{eq:our_F}.

\begin{align}
F = \underbrace{D_{KL}[q(z_t)||p(z_t)]}_{\text{Complexity}} - \underbrace{\mathbb{E}_{q(z_t)}[\log p(o_{t+1}|z_t)]}_{\text{Accuracy}}
\label{eq:our_F}
\end{align}

\noindent
In practice, the complexity term is multiplied by a non-negative hyperparameter $\beta$ describing its relative importance to accuracy. 

Notice that the curiosity term in Equation \ref{eq:all_intrinsic} and the complexity term in Equation \ref{eq:free_energy} are both KL divergences that quantify information gains (a.k.a., relative entropies). Inferring latent states or parameters---by minimizing variational free energy---minimizes the divergence between the posterior and prior, given an observation; thereby minimizing complexity and implicitly maximizing compression and generalization. Conversely, during action selection, based upon expected free energy, the divergence between the predictive posterior and prior is maximized, to maximize expected information gain. One can leverage these information theoretic interpretations by using the complexity term as an estimate of the expected information gain, under the policy being learned. This is licensed because the actor-critic model used in reinforcement learning learns a state-action policy that can, effectively, learn the expected information gain in the same way that it learns expected extrinsic rewards. In other words, one can define hidden state curiosity at the previous time step to be the complexity in Equation \ref{eq:out_F}. (In practice, we apply a clamp constraining hidden state curiosity between 0 and 1). Thus, the actor and critics are trained to minimize expected free energy, rewriting equation \ref{eq:all_intrinsic} as 

\begin{align}
G(s_t, a_t) &=  -\underbrace{D_{KL}[q(z_{t+1})||p(z_{t+1})]}_{\text{Hidden State Curiosity}} - \underbrace{r(s_t, a_t)}_{\text{Extrinsic Reward}} \nonumber \\
&- \underbrace{\mathcal{H}(\pi_\phi(a_t|o_t))}_{\text{Entropy}} - \underbrace{\mathbb{E}_{\pi_\phi(a_t|o_t)} [\log p(a_t|o_t)]}_{\text{Imitation}}. 
\label{eq:out_F}
\end{align}

\noindent
Equations \ref{eq:actor_with_entropy} and \ref{eq:target_value} can incorporate approximations of these entropy and curiosity values, multiplied by non-negative hyperparameters $\alpha$ and $\eta$ (not $\beta$) respectively. This underwrites an adversarial relationship: the forward model is trained to make accurate predictions with minimal complexity but the critic rewards the actor for choosing actions that result in high complexity; namely, a high information gain that characterizes curious or information seeking policies. This is in contrast to prediction error curiosity, which rewards the actor for choosing actions resulting in poor accuracy. No agent can perfectly predict an observation which includes random noise, leaving accuracy perpetually flawed, so agents with prediction error curiosity will always be curious about such an observation. However, an agent employing hidden state curiosity may generate a prior inner state which foresees the observation's random noise; thus, the random noise might not provide any additional information for the posterior inner state, so the agent's prior and posterior conceptualization of the environment can perfectly match despite that randomness. Therefore, rather than seeking merely noisy observations, such agents seek observations which actually alter their latent interpretations of the environment. This reflects how Pathak et al. utilized latent spaces ignoring irrelevant noise, so we predict hidden state curiosity can effectively explore without negative influence from curiosity traps. Also, $p(z_t)$ and $q(z_t)$ can be newly generated for each individual transition in a batch of episodes with relatively easy computational cost, assigning curiosity values in a scalable manner. 

\section{Simulation Experiments}
\label{chap:experiments}

As established in section \ref{chap:introduction}, we designed our experiments to investigate these two hypotheses:

\begin{enumerate}
\item Entropy and curiosity improve agent exploration, especially when both are implemented together as implied by the FEP. 
\item Prediction error curiosity can be negatively influenced by observational noise also known as curiosity traps, while hidden state curiosity can be more resilient to such curiosity traps.
\end{enumerate}

\noindent
To this end, our experiments feature six types of agents training to find goals in various mazes. These will be baseline agents devoid of intrinsic rewards, agents motivated by either entropy or one form of curiosity (prediction error or hidden state), and agents motivated by a combination of entropy and one type of curiosity. See table \ref{tab:hyperparameters} for details about these six types. All agents will share the same architecture with $\gamma = .9$, $\tau = .1$, $d=2$, $\bar{\mathcal{H}} = -1$ (see section \ref{sec:RL}), $\beta = .03$ (see equation \ref{eq:our_F}), and learning rate $\lambda = .01$ with Adam optimizers. No agents have periods of forced investigation with random actions, highlighting the importance of motivating exploration. See tables \ref{tab:forward_model_1} through \ref{tab:critic_model_2} in appendix B for details about models' architectures in PyTorch. Each table illustrates the parameters of a model layer by layer. 

\begin{table}[H]
\centering
\begin{tabular}{|c|c|c|c|c|c|c|}
\hline
Name (and Acronym) & $\alpha$ & $\eta$ & $P_t$\\
\hline
\hline
No Entropy, No Curiosity (N) & 0 & 0 & None\\
\hline
Entropy (E) & None & 0 & None\\
\hline
Prediction Error Curiosity (P) & 0 & 1 & $\mathbb{E}_{q(z_t)}[\log p(o_{t+1}|z_t)]$\\
\hline
Entropy and Prediction Error Curiosity (EP) & None & 1 & $\mathbb{E}_{q(z_t)}[\log p(o_{t+1}|z_t)]$\\
\hline
Hidden State Curiosity (H) & 0 & 1 & $D_{KL}[q(z_{t+1})||p(z_{t+1})]$\\
\hline
Entropy and Hidden State Curiosity (EH) & None & 1 & $D_{KL}[q(z_{t+1})||p(z_{t+1})]$\\
\hline
\end{tabular}
\caption{Hyperparameters for six types of agents. Recall $\alpha$ and $\eta$ in equations \ref{eq:actor_with_entropy} and \ref{eq:target_value}. If $\alpha =$ None, then $\alpha$ is a dynamic parameter chosen to minimize equation \ref{eq:alpha}. In Prediction Error curiosity (P and EP), the curiosity value $P_t$ is the accuracy term of equation \ref{eq:our_F}. In Hidden State Curiosity (H and EH), $P_t$ is the curiosity term of equation \ref{eq:out_F}.}
\label{tab:hyperparameters}
\end{table}

Regarding the first hypothesis, we predict the baseline agent will perform the least efficient exploration, while agents rewarded for both entropy and curiosity will outperform the rest, regardless of which kind of curiosity. However, regarding the second hypothesis, we predict that if we train agents in mazes with curiosity traps, agents with prediction error curiosity will be attracted to those traps, showcasing its susceptibility, while agents with hidden state curiosity are able to ignore them.

\subsection{Experiment Design}

In these experiments, we employ the PyBullet physics engine to simulate an RL agent embodied as a duck. The agent's observations have two parts: its current speed and an 8 by 8 by 4 image of what is in front of it with the four channels being red, green, blue, and distance. The agent's actions also have two parts: adjusting its yaw up to 90 degrees left or right and choosing a speed between 0 meters per time step and a speed limit (with the blocks constructing the mazes having side-length of one meter). 

\begin{figure}[ht!]
    \centering
    \includegraphics[width=0.5\textwidth]{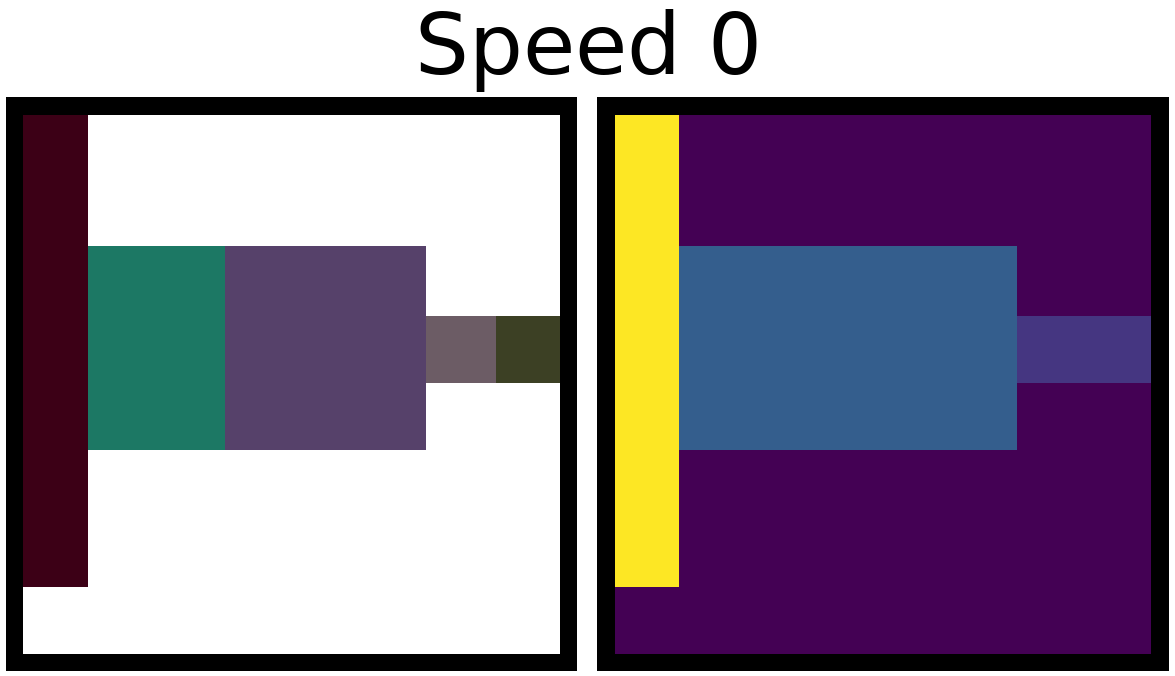}
    \label{fig:observation}

    \caption{An agent's observation includes its current speed in meters per time step and an 8 by 8 by 4 image of what is in front of it. The image's four channels are red, green, blue (left), and distance (right). This is the agent's first observation in the biased T-maze; see figure \ref{fig:tmaze}.}
\end{figure}

Each simulated episode is terminated when the agent exits the maze; the agent's choice of exit will earn an extrinsic reward or punishment. If no exit is chosen within 30 steps, the episode ends with a punishment of $r=-1$. Colliding with a maze wall at any step will also punish the agent with $r=-1$. Any positive extrinsic rewards are multiplied by $.99^{\text{steps taken}}$, encouraging haste. 

In each epoch, the agent will carry out one episode. Memory of that episode's transitions will be saved in that agent's recurrent replay buffer; if that replay buffer contains memory of more than 250 episodes, the oldest episode will be deleted. Then a batch of 32 episodes will be sampled from the replay buffer to train the agent's forward model, actor, and critics. In this manner, we will use different random seeds to train 360 agents of each of the six types described above. This will be carried out both with and without implementing curiosity traps by randomly changing colors of walls near inferior exits with every step, investigating which types of agents are disadvantaged by observational noise.

\begin{figure}[H]
  \centering
  
  \begin{subfigure}[b]{.6\textwidth}
  \centering
  \includegraphics[width=\textwidth]{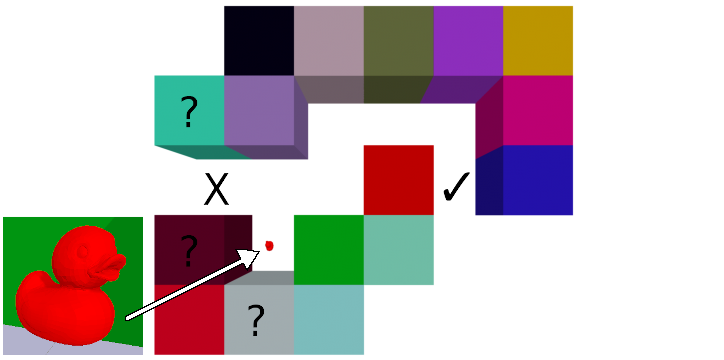}
  \caption{Biased T-Maze.}
  \label{fig:tmaze}
  \end{subfigure}
  
  \vspace{.5cm} 
  
  \begin{subfigure}[b]{0.281\textwidth}
    \centering
    \includegraphics[width=\textwidth]{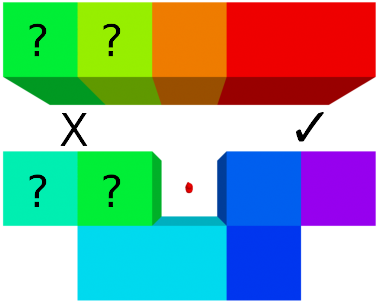}
    \caption{T-Maze.}
    \label{fig:tmaze_1}
  \end{subfigure}
  \hspace{1cm}
  \begin{subfigure}[b]{0.39\textwidth}
    \centering
    \includegraphics[width=\textwidth]{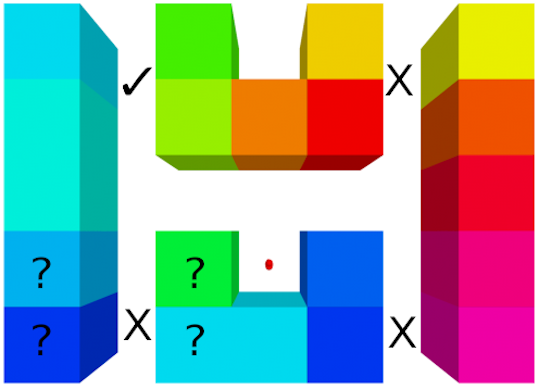}
    \caption{Double T-Maze.}
    \label{fig:tmaze_2}
  \end{subfigure}
  
  \vspace{.5cm} 
  
  \begin{subfigure}[b]{\textwidth}
    \centering
    \includegraphics[width=.493\textwidth]{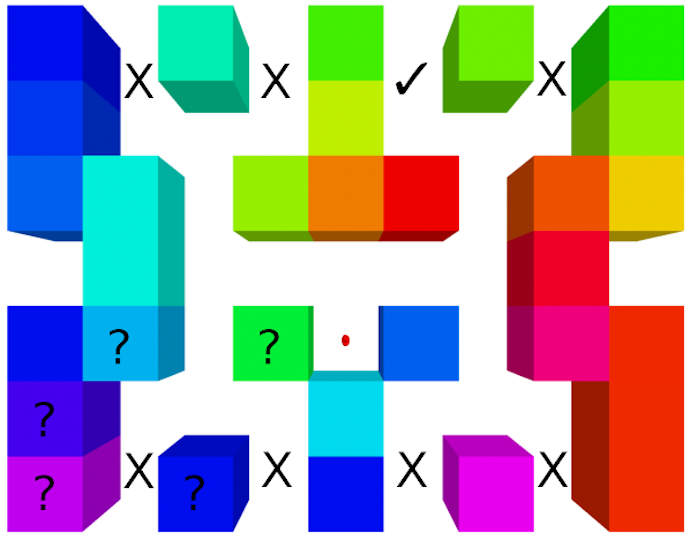}
    \caption{Triple T-Maze.}
    \label{fig:tmaze_3}
  \end{subfigure}
  
  \caption{Agent starts where shown. Correct and incorrect exits are marked \ding{51} and \textsf{\textbf{X}}. With curiosity traps, blocks with \textbf{?} change colors each step. Experiment one uses (a) biased T-Maze. Experiment two uses (b) T-maze, (c) double T-maze, (d) triple T-maze.}
  \label{fig:many_mazes}
\end{figure}

\subsubsection{Biased T-Maze}
\label{sec:t_maze}

In the biased T-maze simulation seen in figure \ref{fig:tmaze}, agents have a speed limit of one meter per time step. The biased T-maze has two exits. The exit out of the T's left arm is easily accessible (nearby and unobstructed) and provides an extrinsic reward of $r=1$. The exit out of the T's right arm is difficult to access (farther away and behind an obstacle) and provides an inconsistent extrinsic reward which is equally likely to be $r=0$ or $r=10$ with an expected value $\mathbb{E}(r) = 5$. Intuitively for human readers, the option with highest expected extrinsic value is the exit to the right despite its distance, so we will call this the correct exit. An agent, however, must explore to the right despite readily available reward on the left to discover the higher value of the correct exit. We trained agents for 500 epochs using the six sets of hyperparameters described in table \ref{tab:hyperparameters} with or without curiosity traps near the incorrect exit.

We predict agents will discover and exploit the higher value of the correct exit more often when encouraged with entropy or curiosity, especially both at once. We also expect curiosity traps to negatively impact performance of agents with prediction error curiosity, while agents with hidden state curiosity are able to ignore them.

\subsubsection{Expanding T-Maze}

In the expanding T-maze simulation, agents have a speed limit of two meters per time step, first in the T-maze seen in figure \ref{fig:tmaze_1}, then the double T-maze seen in figure \ref{fig:tmaze_2}, and then the triple T-maze seen in figure \ref{fig:tmaze_3}. In each of the three mazes, only one exit is deemed correct, with its location alternating between successive mazes to challenge the agents' ability to override previously learned habitual behaviors. If the agent takes the correct exit, it will be rewarded with $r=10$, but if the agent takes any other exit, it will be punished with $r=-.5$. We trained agents for 500 epochs in the T-maze, then for 2000 epochs in the double T-maze, and then for 4000 epochs in the triple T-maze. We trained agents using the six sets of hyperparameters described in table \ref{tab:hyperparameters} with or without curiosity traps near the incorrect exits in the mazes' bottom left portions. 

We predict all agents, even those without intrinsic rewards for exploration, to easily discover and exploit the correct exit on the right side of the T-maze. Then, when relocated to the double T-maze, we predict all agents to first move to the right, away from the correct exit now on the left side. We predict agents without intrinsic rewards for exploration will have difficulty extinguishing that learned behavior, while agents intrinsically rewarded with entropy or curiosity, especially both, are able to begin exploring again to find the new correct exit. Finally, relocation to the triple T-maze will present this challenge again on a larger scale. We also predict curiosity traps to negatively impact performance of agents with prediction error curiosity, while agents with hidden state curiosity are able to ignore them.

\subsection{Results}

\begin{landscape}

\begin{figure}[H]
    \centering
    \includegraphics[width=1.15\textwidth]{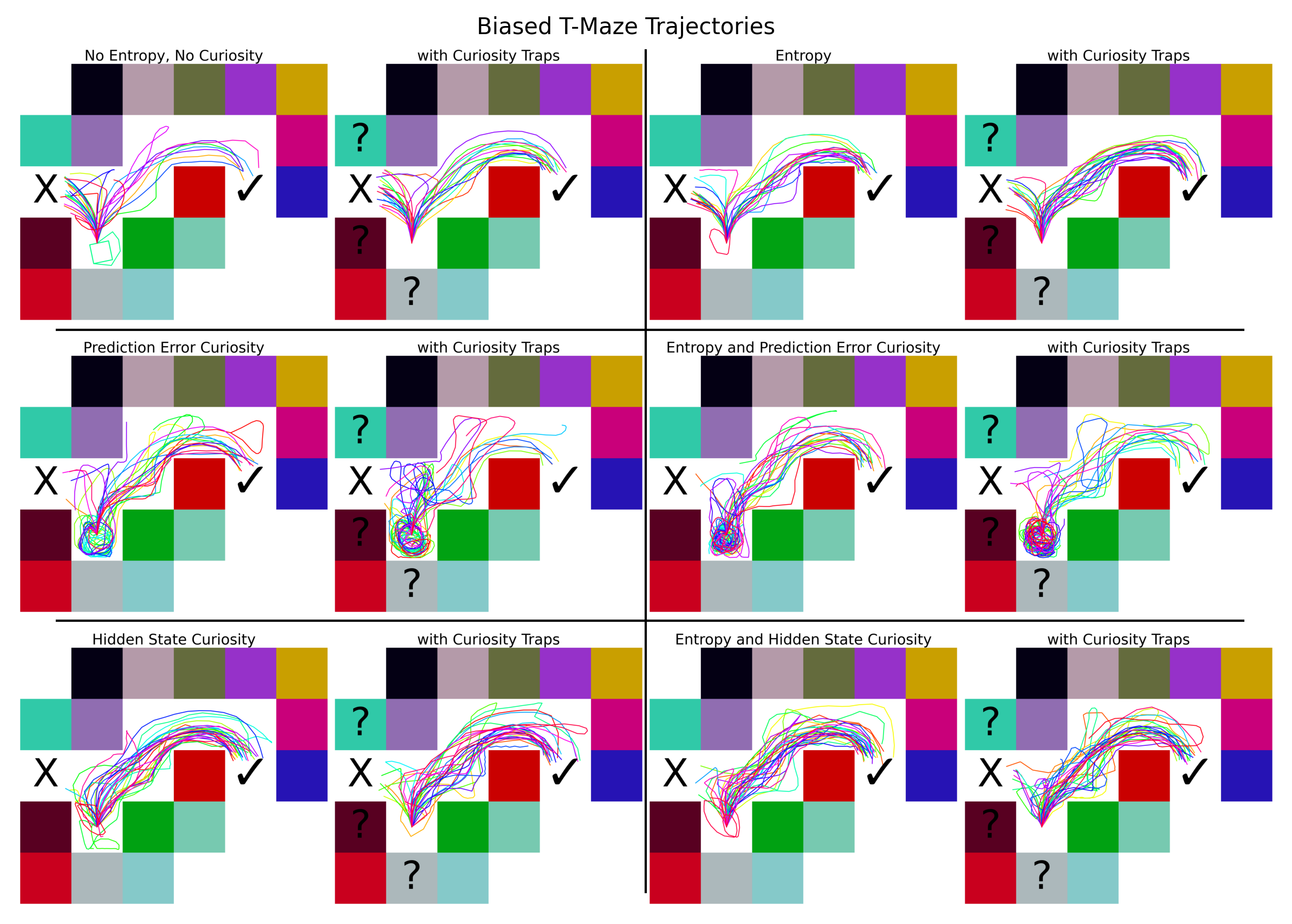}
    \caption{Trajectories of agents after training in the biased T-maze. The correct exit of each maze is marked with a \ding{51}, while each incorrect exit is marked with an \textsf{\textbf{X}}. If curiosity traps are applied, blocks marked with a \textbf{?} will change to random colors with every step.}
    \label{fig:t_maze_paths}
\end{figure}

\end{landscape}

\begin{figure}[H]
    \centering
    \includegraphics[width=.93\textwidth]{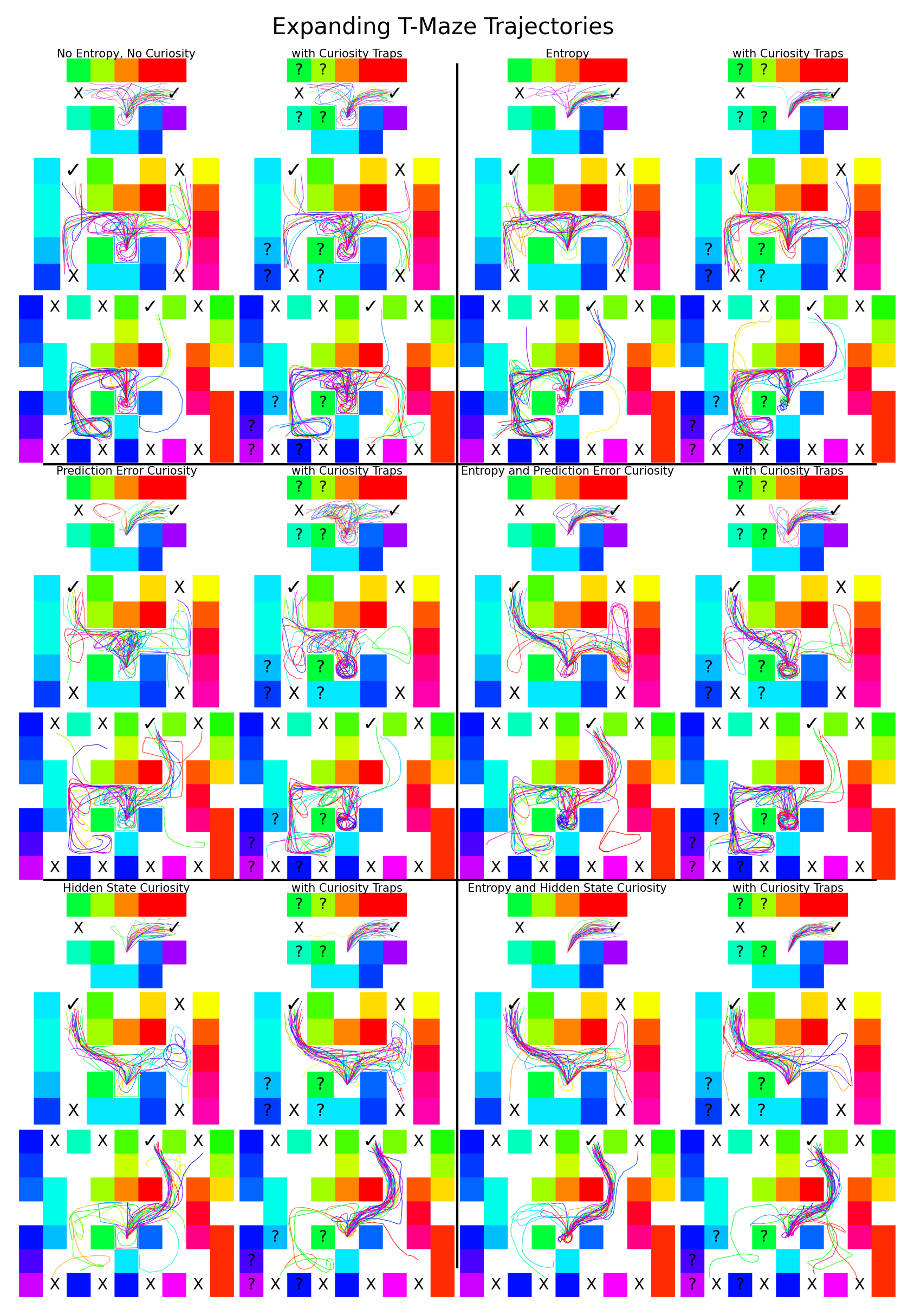}
    \caption{Trajectories of agents after training in the expanding T-maze. The correct exits are marked with a \ding{51}, while incorrect exits are marked with an \textsf{\textbf{X}}. If curiosity traps are applied, blocks marked with a \textbf{?} will change to random colors with every step.}
    \label{fig:many_paths}
\end{figure}

In figures \ref{fig:t_maze_paths} and \ref{fig:many_paths}, see the trajectories of agents trained in the biased T-maze and expanding T-maze, depicting their behaviors. Find an example of an agent's forward model predicting observations in the biased T-maze in figure \ref{fig:predictions}. In figures \ref{fig:t_maze_exits} and \ref{fig:many_exits}, in appendix A, see the proportions of agents choosing each exit in each epoch. Find videos of how agent trajectories changed over time at github.com/oist-cnru/curious\_maze.

\begin{figure}[H]
    \centering
    \includegraphics[width=1\textwidth]{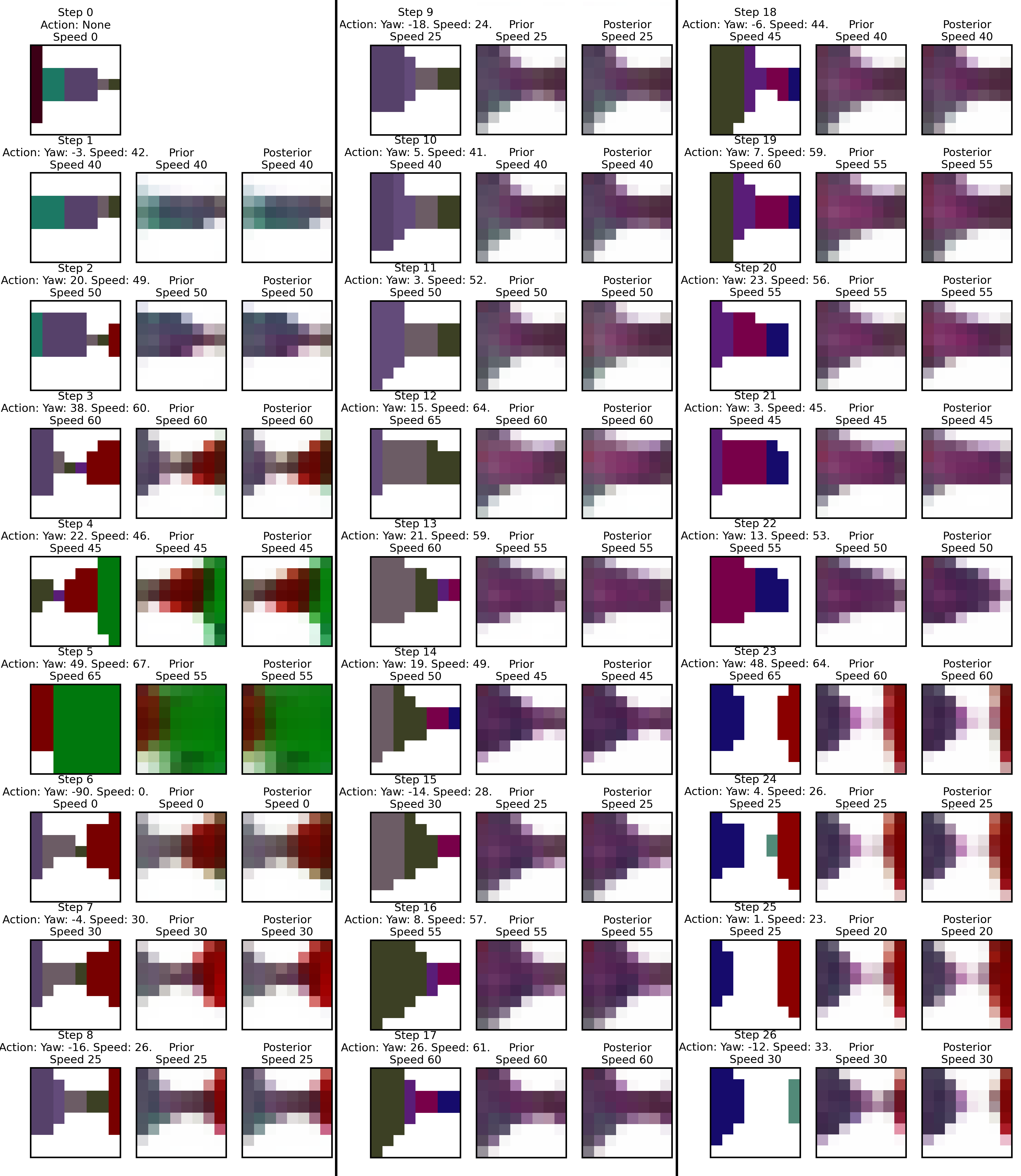}
    \caption{Predictions of agent choosing correct exit in biased T-maze, trained with EH (see table \ref{tab:hyperparameters}) without curiosity traps. Left column: actual observations. Middle column: predictions based on hidden-state $h^p$, made with prior inner state $z^p$. Right column: predictions based on hidden-state $h^q$, made with posterior inner state $z^q$.}
    \label{fig:predictions}
\end{figure}

\begin{figure}[p]
    \centering
    \begin{subfigure}[b]{\textwidth}
        \centering
        \includegraphics[width=.52\textwidth]{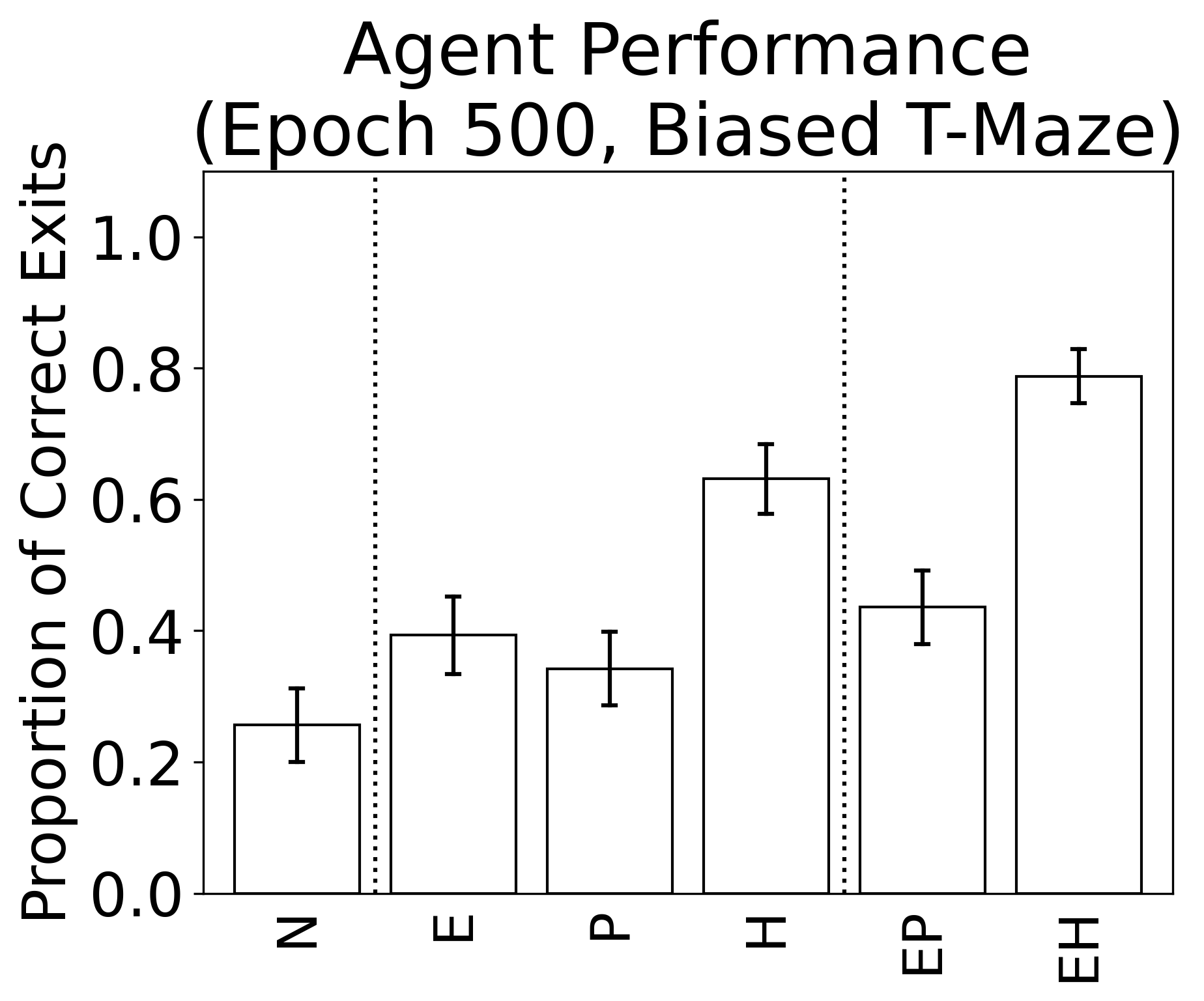}
        \caption{}
        \label{fig:t_maze_hyp_1}
    \end{subfigure}
    
    \vspace{.5cm}
    
    \begin{subfigure}[b]{\textwidth}
        \centering
        \includegraphics[width=.52\textwidth]{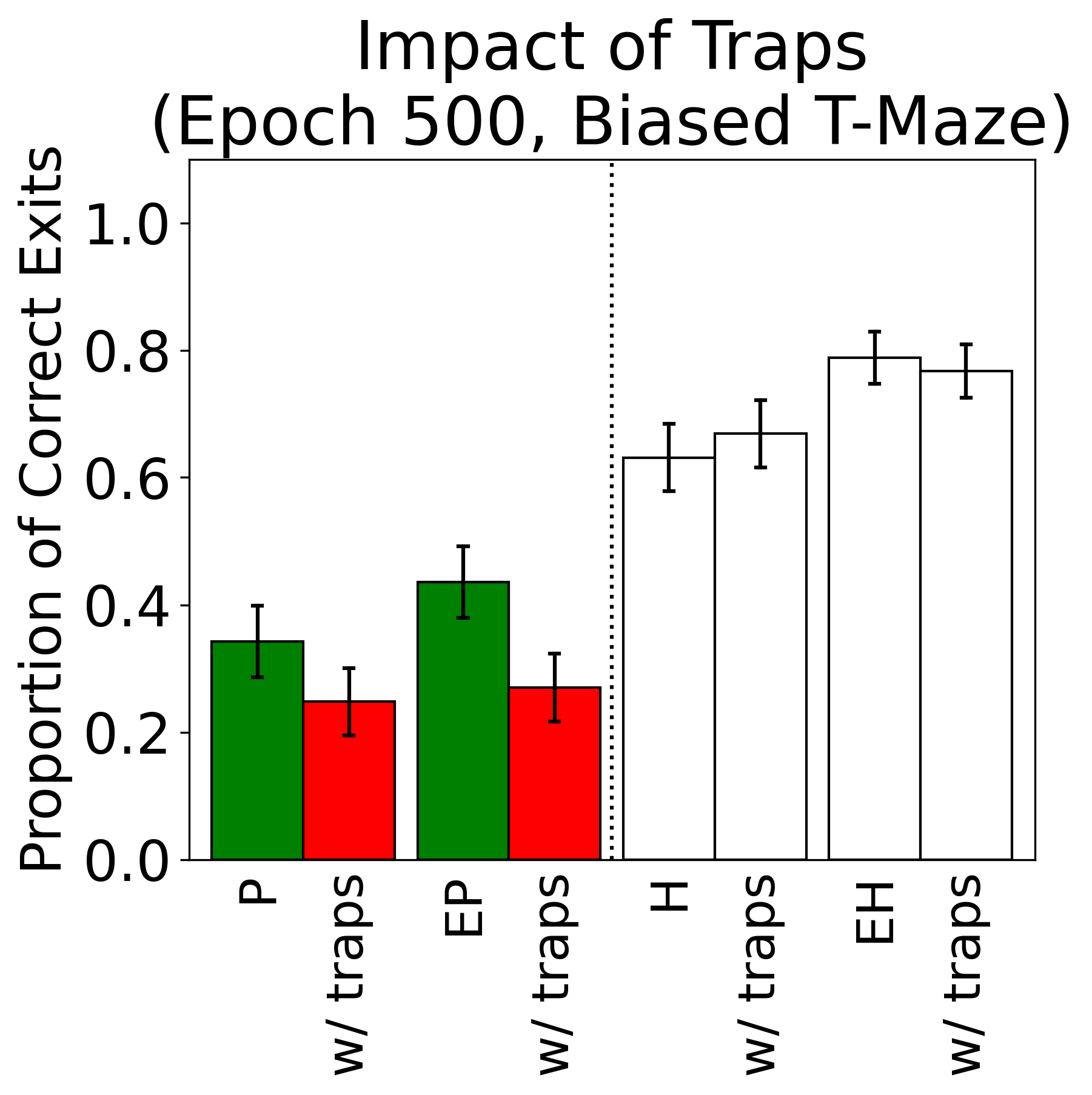}
        \caption{}
        \label{fig:t_maze_hyp_2}
    \end{subfigure}
    
    \caption{Biased T-Maze Results. (a) Bars show rate agents chose the correct exit in their last ten episodes after training using hyperparameters labeled with acronyms from table \ref{tab:hyperparameters}. Error bars show 99\% confidence interval. (b) In each pair of bars, if the left bar is taller than the right bar with confidence of 99\%, the left bar is green and the right bar is red, showing negative impact of curiosity traps.}    
    \label{fig:t_maze_p_value}
\end{figure}

Statistic results of these experiments are displayed in figures \ref{fig:t_maze_p_value} and \ref{fig:many_p_value}, which show how often agents trained with the hyperparameters described in table \ref{tab:hyperparameters} reached the correct exit in their final ten episodes. 

Figures \ref{fig:t_maze_hyp_1} and \ref{fig:many_hyp_1} regard our first hypothesis: entropy and curiosity improve agent exploration, especially when both are implemented together as implied by the FEP. 

Figures \ref{fig:t_maze_hyp_2} and \ref{fig:many_hyp_2} regard our second hypothesis: prediction error curiosity can be negatively influenced by observational noise also known as curiosity traps, while hidden state curiosity can be more resilient to such curiosity traps. 

\subsubsection{Biased T-Maze}

In figure \ref{fig:t_maze_hyp_1}, the leftmost bar shows that agents trained using the hyperparameters labeled ``No Entropy, No Curiosity'' (acronym N) were the least successful agents in the biased T-maze. The three bars between the dotted lines show that agents trained using ``Entropy'' (E), ``Prediction Error Curiosity'' (P), or ``Hidden State Curiosity'' (H) all performed as well as or better than N utilizing one intrinsic reward. The rightmost bars show agents trained with two intrinsic rewards using ``Entropy and Prediction Error Curiosity'' (EP) or ``Entropy and Hidden State Curiosity'' (EH) performed best of all, demonstrating the importance of combining these intrinsic rewards. 

In figure \ref{fig:t_maze_hyp_2}, to the left of the dotted line, note that agents trained using P or EP performed significantly worse when trained with curiosity traps. In contrast, right of the dotted line, agents trained using H or EH have no negative impact from curiosity traps. This demonstrates that hidden state curiosity can mitigate pitfalls which can entrap prediction error curiosity.

These results can be visually confirmed in figures \ref{fig:t_maze_paths} and \ref{fig:t_maze_exits}. In figure \ref{fig:t_maze_paths}, some agents trained using EP with curiosity traps only travel in circles to fixate on the randomly changing walls, revealing the distraction caused by observational noise. Figure \ref{fig:t_maze_exits}, in appendix A, shows how agents trained using N commit to the first exit they encounter, as the rates of exit-choice increase epoch to epoch but never decrease. In contrast, other agents select the incorrect exit at an increasing rate until a peak at approximately the $100^\textrm{th}$ epoch, at which point selection of the correct exit increases instead, as if the agents became bored with the easy exit and explored instead. 

\subsubsection{Expanding T-Maze}

\begin{landscape}

\begin{figure}[p]
    \centering
    \begin{subfigure}[b]{1.35\textwidth}
        \centering
        \includegraphics[width=0.28\textwidth]{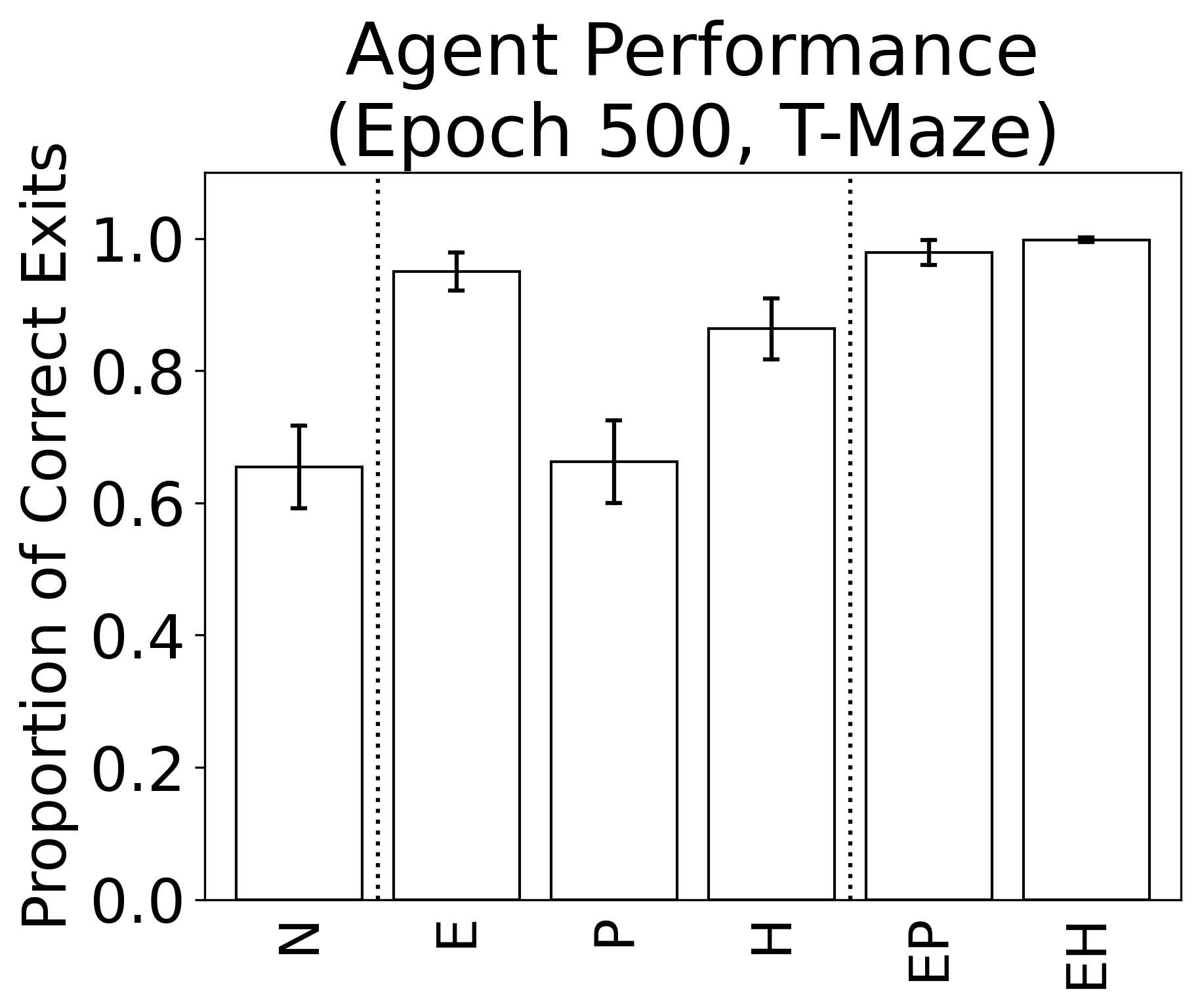}
        \includegraphics[width=0.29\textwidth]{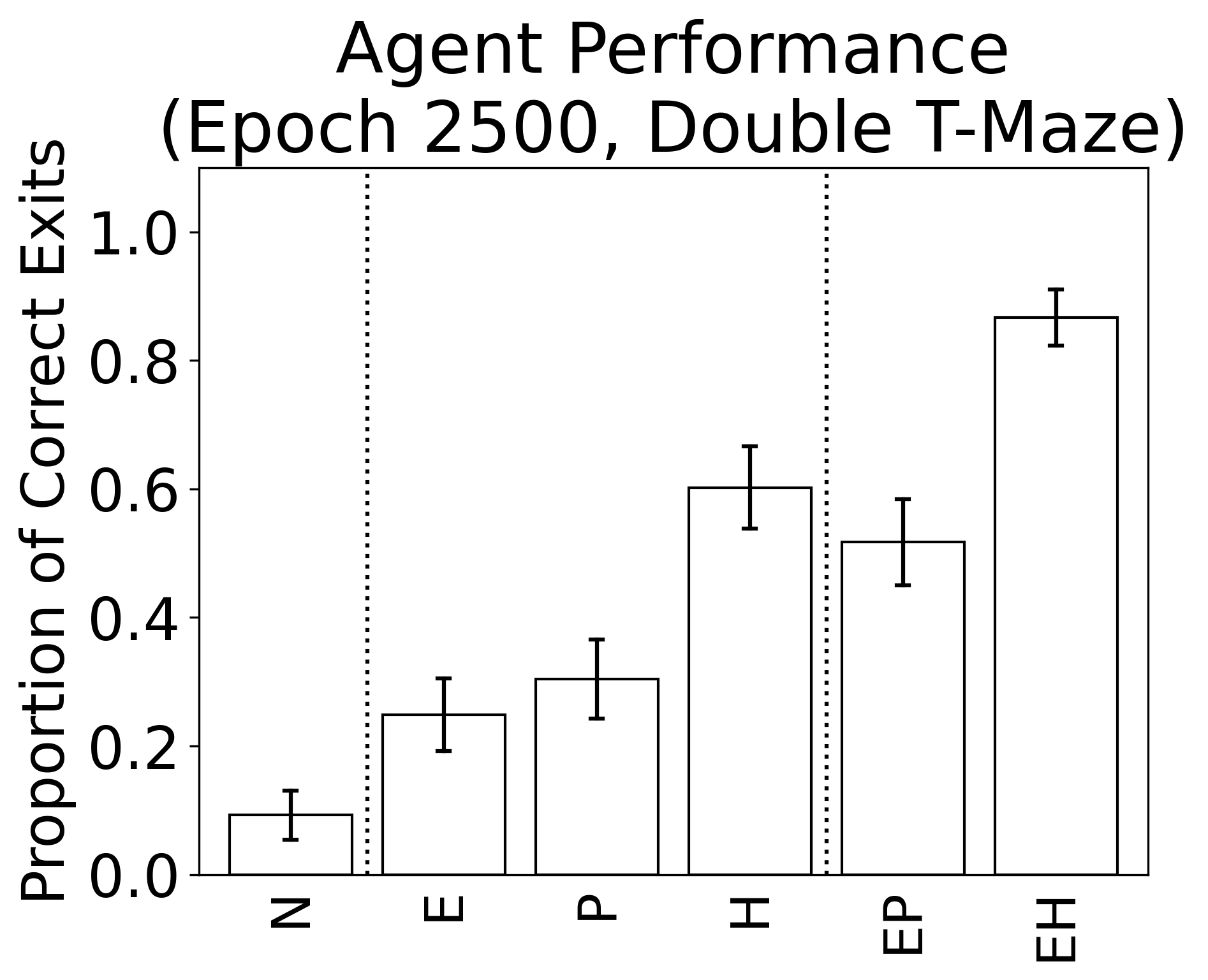}
        \includegraphics[width=0.29\textwidth]{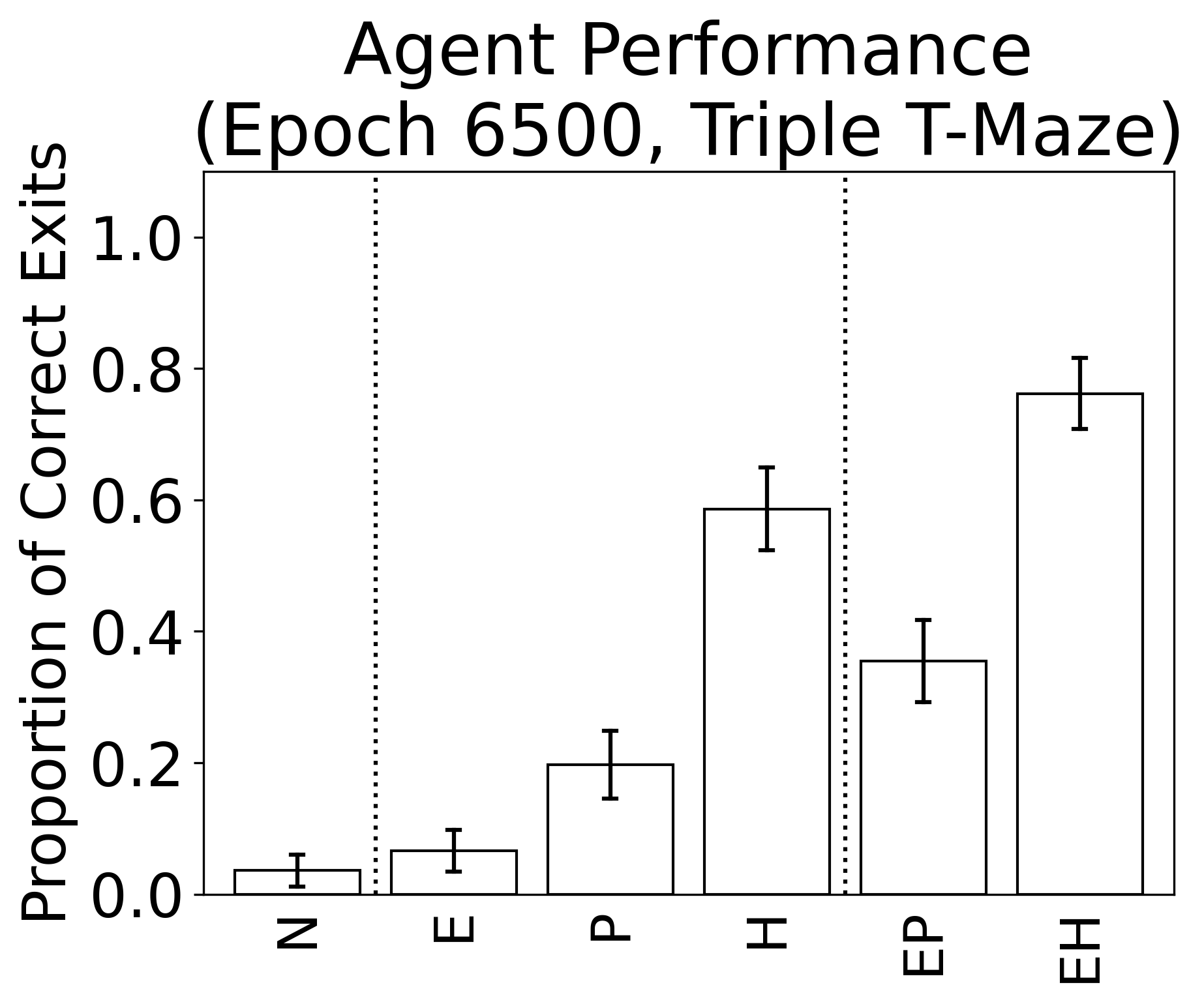}
        \caption{}
        \label{fig:many_hyp_1}
    \end{subfigure}

    \vspace{0.001cm}

    \begin{subfigure}[b]{1.4\textwidth}
        \centering
        \includegraphics[width=0.27\textwidth]{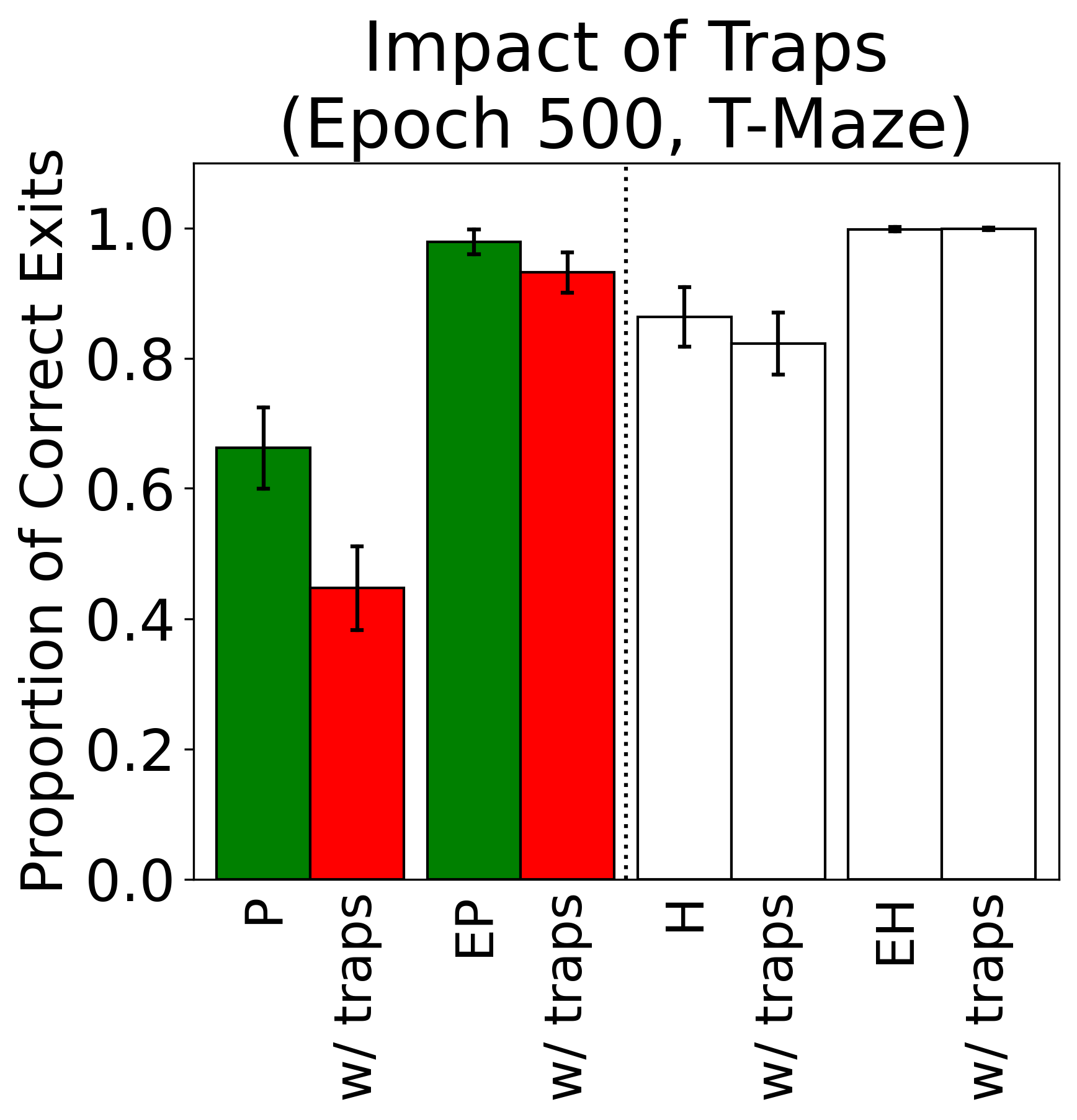}
        \includegraphics[width=0.29\textwidth]{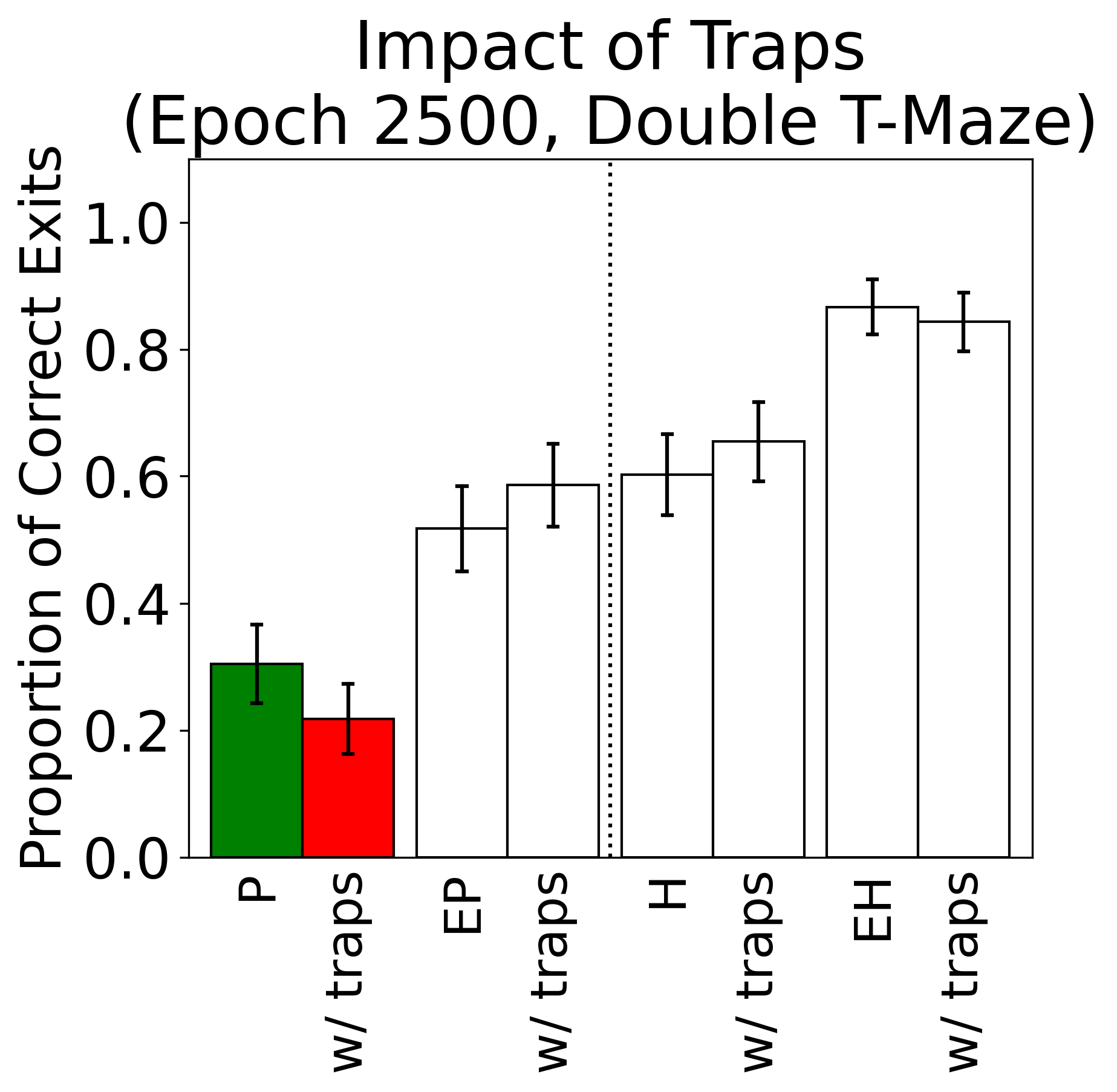}
        \includegraphics[width=0.29\textwidth]{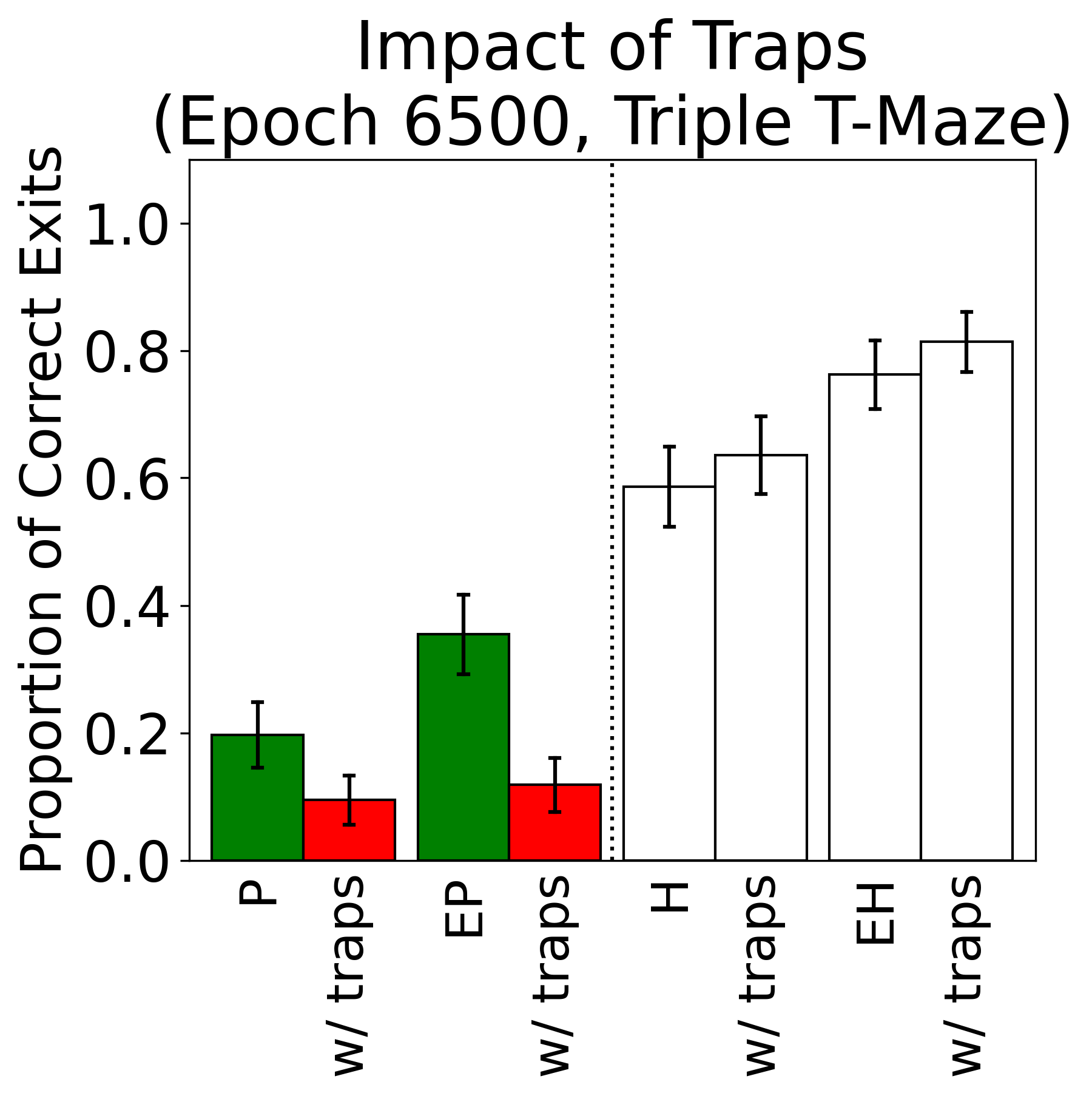}
        \caption{}
        \label{fig:many_hyp_2}
    \end{subfigure}
    
    \caption{Expanding T-Maze Results (see descriptions in figure \ref{fig:t_maze_p_value}).}    
    \label{fig:many_p_value}
\end{figure}

\end{landscape}

Figure \ref{fig:many_hyp_1} displays the performance of agents at the end of training in the T-maze, then the double T-maze, and finally the triple T-maze. Consistent to all three mazes, and just like in the biased T-maze, the leftmost bar shows that agents trained using the hyperparameters labeled ``No Entropy, No Curiosity'' (acronym N) were the least successful agents. The three bars between the dotted lines show that agents trained using ``Entropy'' (E), ``Prediction Error Curiosity'' (P), or ``Hidden State Curiosity'' (H) all performed as well as or better than N with one intrinsic reward. The rightmost bars show agents trained with two intrinsic rewards using ``Entropy and Prediction Error Curiosity'' (EP) or ``Entropy and Hidden State Curiosity'' (EH) performed best of all, demonstrating the importance of combining these intrinsic rewards.

In figure \ref{fig:many_hyp_2}, to the left of the dotted line, note that agents trained using P performed worse when trained with curiosity traps in all three of these mazes, and agents trained using EP were deeply influenced by curiosity traps in the T-maze and triple T-maze. In contrast, right of the dotted line, agents trained using H or EH have no negative impact from curiosity traps. This demonstrates that hidden state curiosity can mitigate pitfalls which can entrap prediction error curiosity.

These results can be visually confirmed in the figures \ref{fig:many_paths} and \ref{fig:many_exits}. In figure \ref{fig:many_paths}, the impact of curiosity traps is clear when some agents trained using P or EP with curiosity traps are attracted to randomly colored walls in all three of the T-mazes. In figure \ref{fig:many_exits}, in appendix A, we see that many agents trained using N learned to reach the correct exit to the right on the T-maze, but many of them continued to select exits on the right side when relocated into the double T-maze even though the correct exit was now on the left. Likewise, many agents trained using E learned to reach the correct exit in the double T-maze with a left turn and then a right turn, but continued to select exits on the left side when relocated into the triple T-maze even through the correct exit was now on the right. In contrast, agents trained using H or EH swiftly stopped choosing incorrect exits when relocated, whether with or without curiosity traps.


\section*{Discussion}

As described in section \ref{chap:experiments}, our experiments corroborated the hypotheses in the introduction: namely, action entropy and curiosity improve agent exploration, especially when both are implemented together as implied by the FEP; prediction error curiosity can be negatively influenced by observational noise also known as curiosity traps, while hidden state curiosity can be more resilient to such curiosity traps. These results indicate that applying the FEP can significantly benefit RL, encouraging agents to investigate and comprehend causal structures which would otherwise be difficult or impossible to understand. This could be beneficial for robots in dynamic environments, or interactive systems automatically personalizing content delivery for its users, or researchers seeking recommendations of directions to survey. However, we have not yet attempted transferring behaviors learned in simulation to physical agents, and only identified the optimized hyperparameters through extensive brute-force testing.

The hidden state curiosity (and action entropy) foregrounded in this work inherit from decompositions of expected free energy. Expected free energy in this setting simply refers to a free energy functional of distributions over (random) variables expected under a particular policy or path into the future. The expected free energy could be regarded as a universal objective function from the perspective of the physics of self organizing agents that have well-defined or characteristic attracting sets. For a recent derivation of expected free energy---from the perspective of statistical physics--please see \cite{friston2023free}. One interesting interpretation of expected free energy is in terms of a dual aspect Bayes optimality. This follows from our decomposition of expected free energy into expected information gain and expected extrinsic reward. These are exactly the objective functions that underwrite the principles of optimal Bayesian experimental design \citep{lindley1956measure, mackay1992practical} and decision theory \citep{berger2011self}, respectively. On this view, intrinsic and extrinsic reward are two sides of the same coin, where information and value have exactly the same currency (i.e., natural units). The implication here is that one can think about rewards in terms of information and, conversely, think about the value of information \citep{howard1966decision} as intrinsically rewarding.

The expected information gain can be applied to any latent variables (i.e. states or parameters) of a forward model. When applied to the latent states of a generative model, the implicit intrinsic reward or motivation is sometimes referred to as salience \citep{itti2009bayesian}. Conversely, when applied to the parameters of a forward model the expected information gain is sometimes referred to as the novelty that underwrites curious behavior \citep{baldassarre2014intrinsic, dacostaActiveInference, schmidhuber2010formal, schwartenbeck2019}. This is important because we have absorbed the implicit active inference and learning into a reinforcement learning scheme based upon the actor-critic model. This kind of reinforcement learning identifies state-action policies in the sense that an optimal policy is identified for every given state. This means one cannot select policies that maximize information gain about latent states (because each policy is conditioned upon being in a particular state; as opposed to having posterior beliefs about latent states). However, it is still possible to learn policies that, on average, are information seeking; especially about the parameters of a generative model---as in our case. The parameters in question here are the transition parameters of the forward model. This leads to the interesting notion that one can learn state-action policies that are information seeking, in exactly the way we have demonstrated with the above numerical experiments. These are sometimes referred to as epistemic habits \citep{friston2016active}: e.g., habitually watching a certain news channel in the evening to seek information about what happened during the day. This observation is potentially important because it suggests there are lawful and learnable ways of foraging changing environments for information; such as mazes that feature curiosity traps and change over time.

Looking forward, there are multiple ways future research can explore hidden state curiosity. For example, the PV-RNN architecture \citep{pvrnn} can introduce hierarchical processing within a VRNN \citep{chung2016recurrent}. Each layer of such a framework could generate hidden states with different Multiple Timescale RNNs (MTRNNs) \citep{yamashita2008, jian2023}, allowing agents to access both long-term and short-term memories. Using this architecture, agents could have curiosity about the environment in multiple temporal contexts. 

Moreover, just as the hyperparameter $\alpha$ can be a parameter which dynamically adjusts to satisfy target entropy $\bar{\mathcal{H}}$ (see section \ref{sec:RL}), it may be possible to adjust the hyperparameter $\eta$ dynamically to satisfy target curiosity $\bar{\mathcal{C}}$. For example, like $\alpha$ is optimized to minimize $\log (\alpha) \cdot (\mathcal{H}(\pi(a_t|o_t)) - \bar{\mathcal{H}})$, $\eta$ could be optimized to minimize $\log (\eta) \cdot (D_{KL}[q(z_t)||p(z_t)] - \bar{\mathcal{C}})$. This may refine the agent's engagement with curiosity over time and help users select optimal hyperparameters.

Also, future research should investigate how choice of $\eta$, $\beta$, and the sizes of observations and inner states impact hidden state curiosity's reaction to curiosity traps. Although choice of $\eta$ and $\beta$ can be considered configurable customization, it can be challenging to fine-tune optimal pairings for ignoring useless noise in the task at hand\footnote{Generally, in FEP-based schemes, a free hyperparameter can, in principle, be optimized with respect to variational free energy. For simple hyperparameters, this is usually best achieved with a line search over the hyperparameter to minimise the path integral of variational free energy---as a bound on log marginal likelihood or model evidence---accumulated over the time period in question.}.

Finally, the intrinsic reward for imitation shown by \citet{kawahara} in equation \ref{eq:all_intrinsic} should be investigated. This could enable agents to learn from human demonstrations, particularly for rare or complex situations.

In conclusion, emulating behaviors associated with biological agents like curiosity-driven exploration appears to be a promising frontier in advancing AI. Although current RL agents can have great computational power, they cannot yet achieve understandings as nuanced as inquisitive humans. The FEP offers a principled way to bring such useful organic practices to artificial agents. Our future hope is to apply hidden state curiosity to more intricate 3D agents training to perform compositional action goals, delving into the FEP's influence on embodied cognition and the emergence of communication.

Find our code at github.com/oist-cnru/curious\_maze.

\newpage
\subsection*{Acknowledgments}
The authors are funded by OIST graduate school.

\noindent
Thank you greatly to the anonymous reviewer.

\bibliographystyle{apalike}
\bibliography{bib.bib}

\clearpage
\begin{landscape}

    \begin{figure}[ht]
        \centering
        \includegraphics[width=1.5\textwidth]{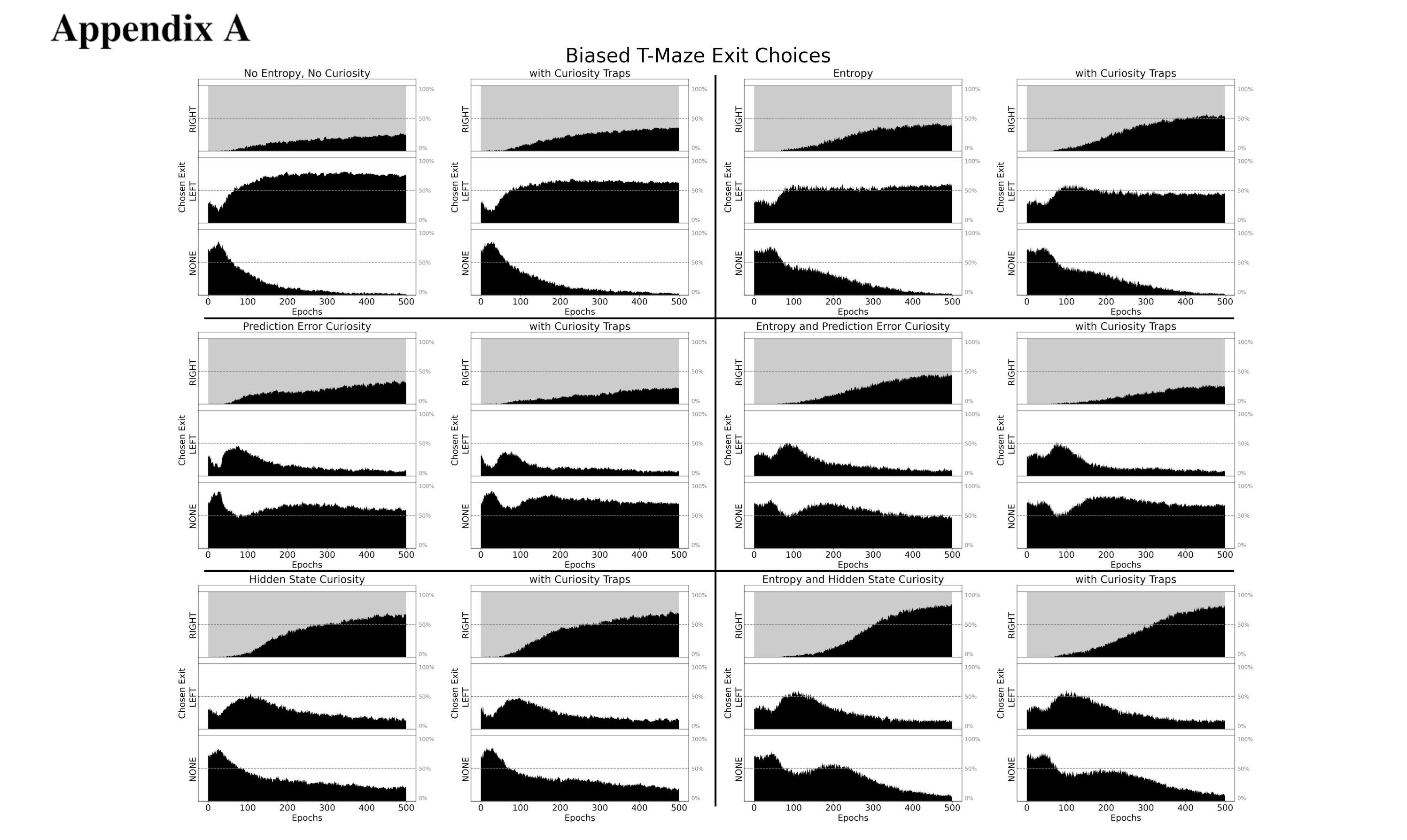}
        \caption{Proportion of agents taking each exit in the T-maze. The correct exit is colored light gray.}
        \label{fig:t_maze_exits}
    \end{figure}

    \begin{figure}[ht]
        \centering
        \includegraphics[width=1.2\textwidth]{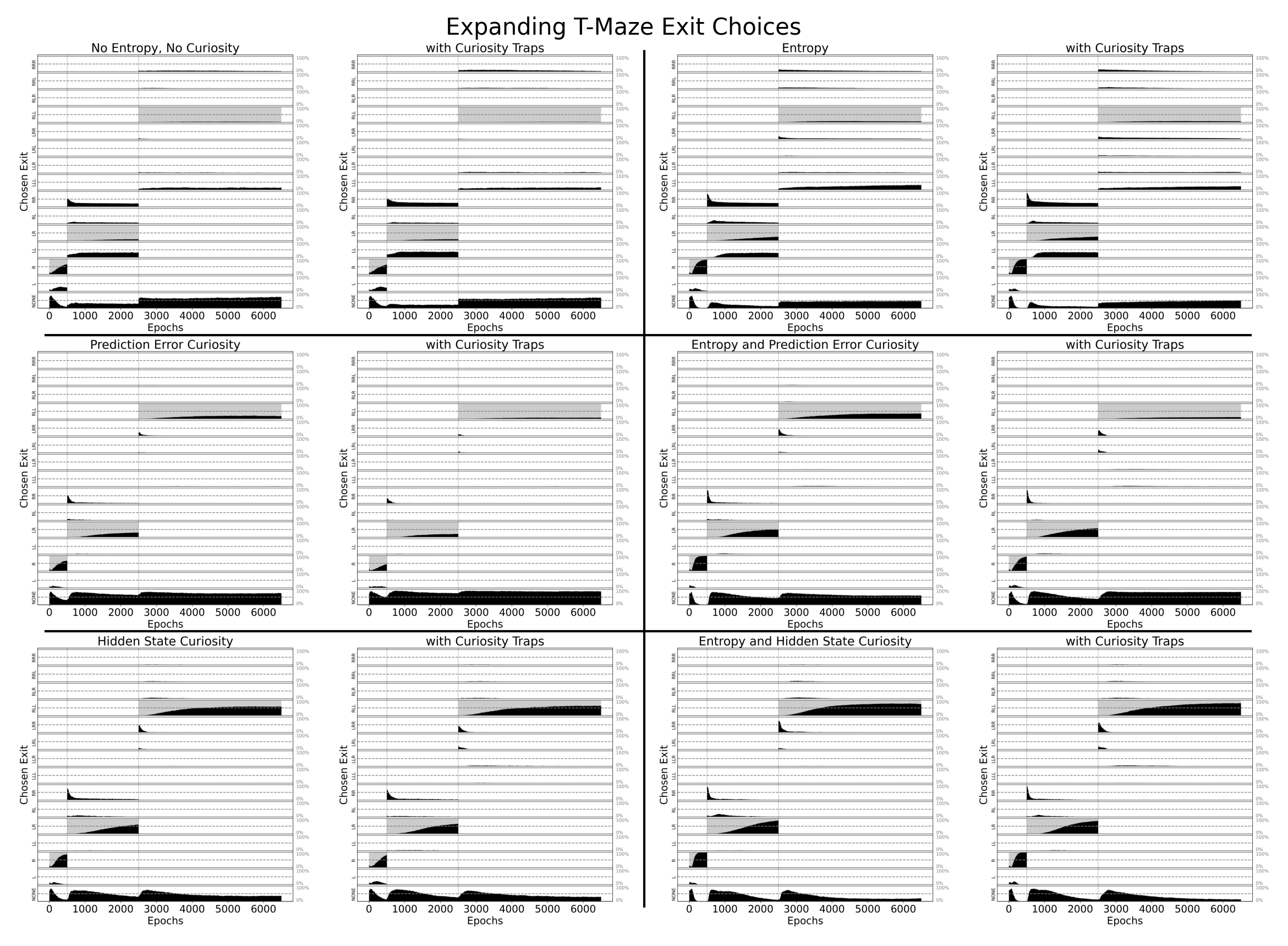}
        \caption{Proportion of agents taking each exit in the expanding T-maze. The correct exit is colored light gray.}
        \label{fig:many_exits}
    \end{figure}
\end{landscape}

\clearpage
\section*{Appendix B}

These tables illustrate details about models' architectures in PyTorch. Each table illustrates the parameters of a model layer by layer. PReLU refers to LeakyReLU with leak-coefficient as a trainable parameter.

\begin{table}[ht]
\centering
\begin{tabular}{llll}
Forward &  &  \\
\hline
Portion & Layer Type & Details \\
\hline
Image In & Convolution & channels in=4, channels out=16\\ 
& & \quad kernel\_size=(3, 3), stride=(1, 1), \\
& & \quad padding=(1, 1), padding\_mode=reflect \\
& PReLU & num\_parameters=1 \\
& Average Pooling & kernel\_size=(3, 3), stride=(2, 2), padding=(1, 1)\\
& Convolution & channels in=16, channels out=16,  \\
& & \quad kernel\_size=(3, 3), stride=(1, 1), \\
& & \quad padding=(1, 1), padding\_mode=reflect \\
& PReLU & num\_parameters=1 \\
& Average Pooling & kernel\_size=(3, 3), stride=(2, 2), padding=(1, 1)\\
& Flatten & \\
& Linear & in\_features=64, out\_features=32, bias=True\\
& PReLU & num\_parameters=1 \\
\hline
Speed In & Linear & in\_features=1, out\_features=32, bias=True\\
& PReLU & num\_parameters=1 \\
\hline
\end{tabular}
\caption{Architecture of the Forward Model (part 1).}
\label{tab:forward_model_1}
\end{table}

\begin{table}[ht]
\centering
\begin{tabular}{llll}
Forward (continued) &  &  \\
\hline
Portion & Layer Type & Details \\
\hline
Action In & Linear & in\_features=2, out\_features=32, bias=True\\
& PReLU & num\_parameters=1 \\
\hline
$\mu^p$ ($z^p$ Mean ) & Linear & in\_features=64, out\_features=32, bias=True\\
& PReLU & num\_parameters=1 \\
& Linear & in\_features=32, out\_features=32, bias=True\\
& Tanh &  \\
\hline
$\sigma^p$ ($z^p$ STD) & Linear & in\_features=64, out\_features=32, bias=True\\
& PReLU & num\_parameters=1 \\
& Linear & in\_features=32, out\_features=32, bias=True\\
& Softplus & beta=1, threshold=20 \\
\hline
$\mu^q$ ($z^q$ Mean) & Linear & in\_features=128, out\_features=32, bias=True\\
& PReLU & num\_parameters=1 \\
& Linear & in\_features=32, out\_features=32, bias=True\\
& Tanh &  \\
\hline
$\sigma^q$ ($z^q$ STD) & Linear & in\_features=128, out\_features=32, bias=True\\
& PReLU & num\_parameters=1 \\
& Linear & in\_features=32, out\_features=32, bias=True\\
& Softplus & beta=1, threshold=20 \\
\hline
\end{tabular}
\caption{Architecture of the Forward Model (part 2).}
\label{tab:forward_model_2}
\end{table}

\begin{table}[ht]
\centering
\begin{tabular}{llll}
Forward (continued) &  &  \\
\hline
Portion & Layer Type & Details \\
\hline
GRU & Gated RNN & input size=32, hidden size=32 \\
\hline
Image Out & Linear & in\_features=64, out\_features=16, bias=True\\
& PReLU & num\_parameters=1 \\
& Reshape \\
& Convolution & channels in=4, channels out=16, \\
& & \quad kernel\_size=(3, 3), stride=(1, 1), \\
& & \quad padding=(1, 1), padding\_mode=reflect \\
& PReLU & num\_parameters=1 \\
& Upsampling &  scale\_factor=2, mode=`bilinear'\\
& Convolution & channels in=16, channels out=16, \\
& & \quad kernel\_size=(3, 3), stride=(1, 1), \\
& & \quad padding=(1, 1), padding\_mode=reflect \\
& PReLU & num\_parameters=1 \\
& Upsampling &  scale\_factor=2, mode=`bilinear'\\
& Convolution & channels in=16, channels out=16,  \\
& & \quad kernel\_size=(3, 3), stride=(1, 1), \\
& & \quad padding=(1, 1), padding\_mode=reflect \\
& PReLU & num\_parameters=1 \\
& Convolution & channels in=16, channels out=4, \\
& & \quad kernel\_size=(1, 1), stride=(1, 1) \\
\hline
\end{tabular}
\caption{Architecture of the Forward Model (part 3).}
\label{tab:forward_model_3}
\end{table}

\begin{table}[ht]
\centering
\begin{tabular}{llll}
Forward (continued) &  &  \\
\hline
Portion & Layer Type & Details \\
\hline
Speed Out & Linear & in\_features=64, out\_features=32, bias=True\\
& PReLU & num\_parameters=1 \\
& Linear & in\_features=32, out\_features=32, bias=True\\
& PReLU & num\_parameters=1 \\
& Linear & in\_features=32, out\_features=1, bias=True\\
\hline
\end{tabular}
\caption{Architecture of the Forward Model (part 4).}
\label{tab:forward_model_4}
\end{table}

\begin{table}[ht]
\centering
\begin{tabular}{llll}
Actor &  &  \\
\hline
Portion & Layer Type & Details \\
\hline
h In & Linear & in\_features=32, out\_features=32, bias=True\\
& PReLU & num\_parameters=1 \\
& Linear & in\_features=32, out\_features=32, bias=True\\
& PReLU & num\_parameters=1 \\
& Linear & in\_features=32, out\_features=32, bias=True\\
& PReLU & num\_parameters=1 \\
& Linear & in\_features=32, out\_features=32, bias=True\\
& PReLU & num\_parameters=1 \\
\hline
$\mu$ & Linear & in\_features=32, out\_features=2, bias=True\\
\hline
$\sigma$ & Linear & in\_features=32, out\_features=2, bias=True\\
& Softplus & beta=1, threshold=20\\
\hline
\end{tabular}
\caption{Architecture of the Actor Model. The final action is $a=\tanh(x \sim \mathcal{N}(\mu,\sigma))$, and the log probability of that action is $\log(\mathcal{N}(x|\mu,\sigma^2))-\log(1-a^2 +10^{-6})$.}
\label{tab:actor_model}
\end{table}

\begin{table}[ht]
\centering
\begin{tabular}{llll}
Critic &  &  \\
\hline
Portion & Layer Type & Details \\
\hline
Image In & Convolution & channels in=4, channels out=16, \\
& & \quad kernel\_size=(3, 3), stride=(1, 1), \\
& & \quad padding=(1, 1), padding\_mode=reflect \\
& PReLU & num\_parameters=1 \\
& Average Pooling & kernel\_size=(3, 3), stride=(2, 2), padding=(1, 1)\\
& Convolution & channels in=16, channels out=16,  \\
& & \quad kernel\_size=(3, 3), stride=(1, 1), \\ 
& & \quad padding=(1, 1), padding\_mode=reflect \\
& PReLU & num\_parameters=1 \\
& Average Pooling & kernel\_size=(3, 3), stride=(2, 2), padding=(1, 1)\\
& Flatten & \\
& Linear & in\_features=64, out\_features=32, bias=True\\
& PReLU & num\_parameters=1 \\
\hline
Speed In & Linear & in\_features=1, out\_features=32, bias=True\\
& PReLU & num\_parameters=1 \\
\hline
Action In & Linear & in\_features=2, out\_features=32, bias=True\\
& PReLU & num\_parameters=1 \\
\hline
\end{tabular}
\caption{Architecture of the Critic Model (part 1). Not pictured: concatenation of Image In, Speed In, and Action In for GRU input.}
\label{tab:critic_model_1}
\end{table}

\begin{table}[ht]
\centering
\begin{tabular}{llll}
Critic (continued) &  &  \\
\hline
Portion & Layer Type & Details \\
\hline
GRU & Gated RNN & input size=96, hidden size=32 \\
\hline
Q Out & Linear & in\_features=32, out\_features=32, bias=True\\
& PReLU & num\_parameters=1 \\
& Linear & in\_features=32, out\_features=1, bias=True\\
\hline
\end{tabular}
\caption{Architecture of the Critic Model (part 2).}
\label{tab:critic_model_2}
\end{table}

\end{document}